\documentclass[10pt,journal]{IEEEtran}
\usepackage{enumitem,calc}
\usepackage{graphicx}
\usepackage{subfigure}
\usepackage{amsmath}
\usepackage{array}
\usepackage{subfig}
\usepackage{tabu}
\usepackage{amssymb}
\usepackage{multirow}
\usepackage{booktabs}
\usepackage{cite}
\usepackage{xcolor}
\usepackage{soul}
\soulregister\cite7
\soulregister\mathbf7

\usepackage[table]{xcolor}

\usepackage{rotating}
\usepackage{array}
\usepackage{makecell}

\usepackage{algorithm}
\usepackage{algorithmic}

\usepackage{multirow}
\usepackage{xcolor}         
\usepackage{bbding}
\usepackage{amssymb}
\usepackage{pifont}
\usepackage{soul}

\usepackage{xcolor}
\usepackage[table]{xcolor}
\usepackage{balance}

\ifCLASSINFOpdf
\else
\fi
\begin{document}

\title{MAC: Masked Agent Collaboration Boosts Large Language Model Medical Decision-Making}

\author{
        Zhihao Peng\textsuperscript{{\dag}},
        Liuxin Bao\textsuperscript{{\dag}},
        Yixuan Yuan

\thanks{{\dag} These authors contributed equally to this work.}
\thanks{This work was supported in part by . \emph{(Corresponding author: Yixuan Yuan.)} }

\thanks{L. Bao is with the School of Automation, Hangzhou Dianzi University, Hangzhou 310018, China (e-mail: lxbao@hdu.edu.cn).}

\thanks{Z. Peng and Y. Yuan are with the Department of Electronic Engineering, Chinese University of Hong Kong, Shatin 000000, China (e-mail: zhihao.peng@cityu.edu.hk; yxyuan@ee.cuhk.edu.hk).}




}

\maketitle

\begin{abstract}
Large language models (LLMs) have proven effective in artificial intelligence, where the multi-agent system (MAS) holds considerable promise for healthcare development by achieving the collaboration of LLMs. However, the absence of a systematic pipeline for agent construction and the rigidity of static collaboration patterns render current MAS-based models vulnerable to collaboration failures, resulting in substantial performance degradation in medical decision-making scenarios. To this end, we propose a novel Masked Agent Collaboration (MAC) framework that harnesses Pareto-optimal agent construction and cross-consistency maximization mechanisms to achieve adaptive progressive propagation of collaborative information, boosting the medical decision-making capacity. Specifically, we first conduct a Pareto-frontier factors analysis towards the LLMs pool to consider their key factors, including the model size, inference time, diversity score, and throughput ratio, where we calculate the similarity between pairwise outputs within an LLM to derive its diversity score.  Beyond this analysis, we enable the identification of Pareto-optimal models that balance efficiency and capability, which are subsequently selected as collaborative agents to consider the fundamental trade-offs inherent in practical LLM deployment. Afterward, we measure the pairwise similarity between the outputs from collaborative agents to determine their cross-consistency values, subsequently masking out the agent with the lowest cross-consistency value to eliminate the output that is likely semantically inconsistent. Finally, we conduct collaboration of agents by achieving adaptive progressive propagation, where each agent aggregates the outputs of unmasked agents from the previous layer as its input to generate the corresponding output via prompt engineering. \textcolor{black}{Evaluations confirm the effectiveness of our MAC, notably outperforming the multi-agent collaboration model (composed of 70B-141B open-access LLMs) by 16.55\% and Claude Haiku 4.5 by 5.29\% on NEJMQA. Our code is available at {github.com/CUHK-AIM-Group/MAC-Masked-Agent-Collaboration}}. 
\end{abstract}

\begin{IEEEkeywords}
Large Language Model, Multi-Agent System, Pareto Frontier, Cross-Consistency Maximization, Adaptive Progressive Propagation, Healthcare
\end{IEEEkeywords}

\IEEEpeerreviewmaketitle

\section{Introduction}

\noindent In recent decades, considerable efforts have been made in developing traditional machine learning approaches and deep learning-based models, enhancing the accuracy and accessibility of medical decision-making healthcare systems \cite{titano2018automated,ardila2019end,tiu2022expert,khajehnejad2025dynamic,mencattini2025mea}. For example, in neuroimaging, an advanced 3D convolutional neural network has been presented to perform rapid and automated screening of non-contrast head computed tomography scans for critical conditions like intracranial hemorrhage and mass effect \cite{titano2018automated}. In lung cancer screening, a deep learning algorithm that analyzes current and prior low-dose computed tomography scans has demonstrated the ability to match or even surpass the performance of radiologists in predicting malignancy risk \cite{ardila2019end}. In chest X-ray interpretation, emerging self-supervised learning models trained on large volumes of unlabeled images have achieved diagnostic accuracy comparable to radiologists for certain pathologies and have shown the ability to generalize to tasks and findings for which they received no explicit expert annotations \cite{tiu2022expert}. Nevertheless, a substantial performance gap remains between the algorithm development and clinical deployment in the healthcare domain. 


\begin{figure}[!]
    \centering
    \includegraphics [width=0.98 \linewidth]{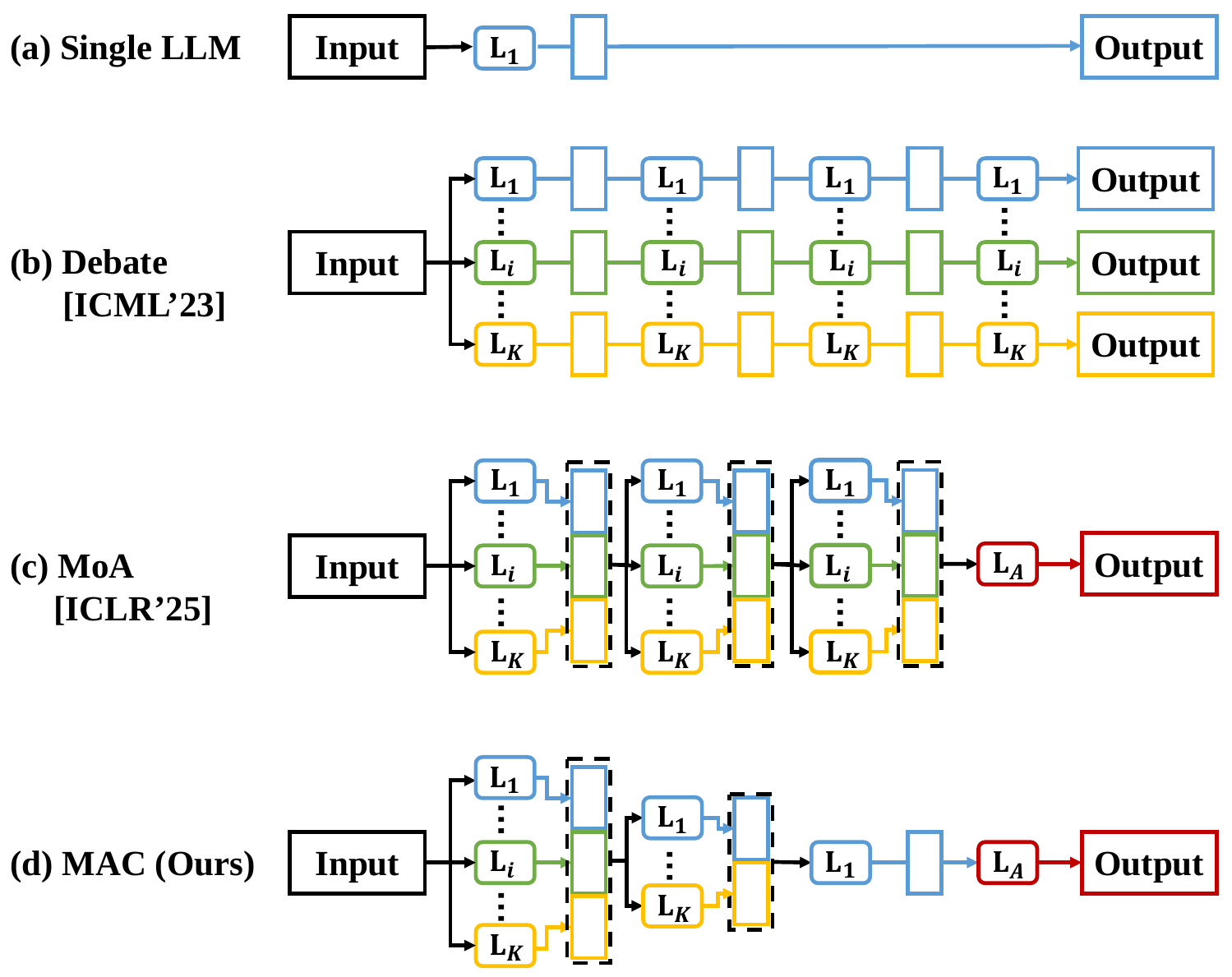}
    \caption{Illustration of (a) single LLM and MAS-based framework architectures, including (b) Debate \cite{estornell2024multi}, (c) MoA \cite{wang2025mixtureofagents}, and (d) MAC (Ours). Different colors represent different agents with responses, where $\textbf{L}_i$ and $\textbf{L}_A$ denote the $i$-th agent and the agent for aggregation, respectively, and $K$ denotes the number of agents in the MAS.}
    \label{fig: Comp_framework}
  \end{figure}


\begin{figure}[t]
    \centering
          \includegraphics[width=0.98\linewidth]{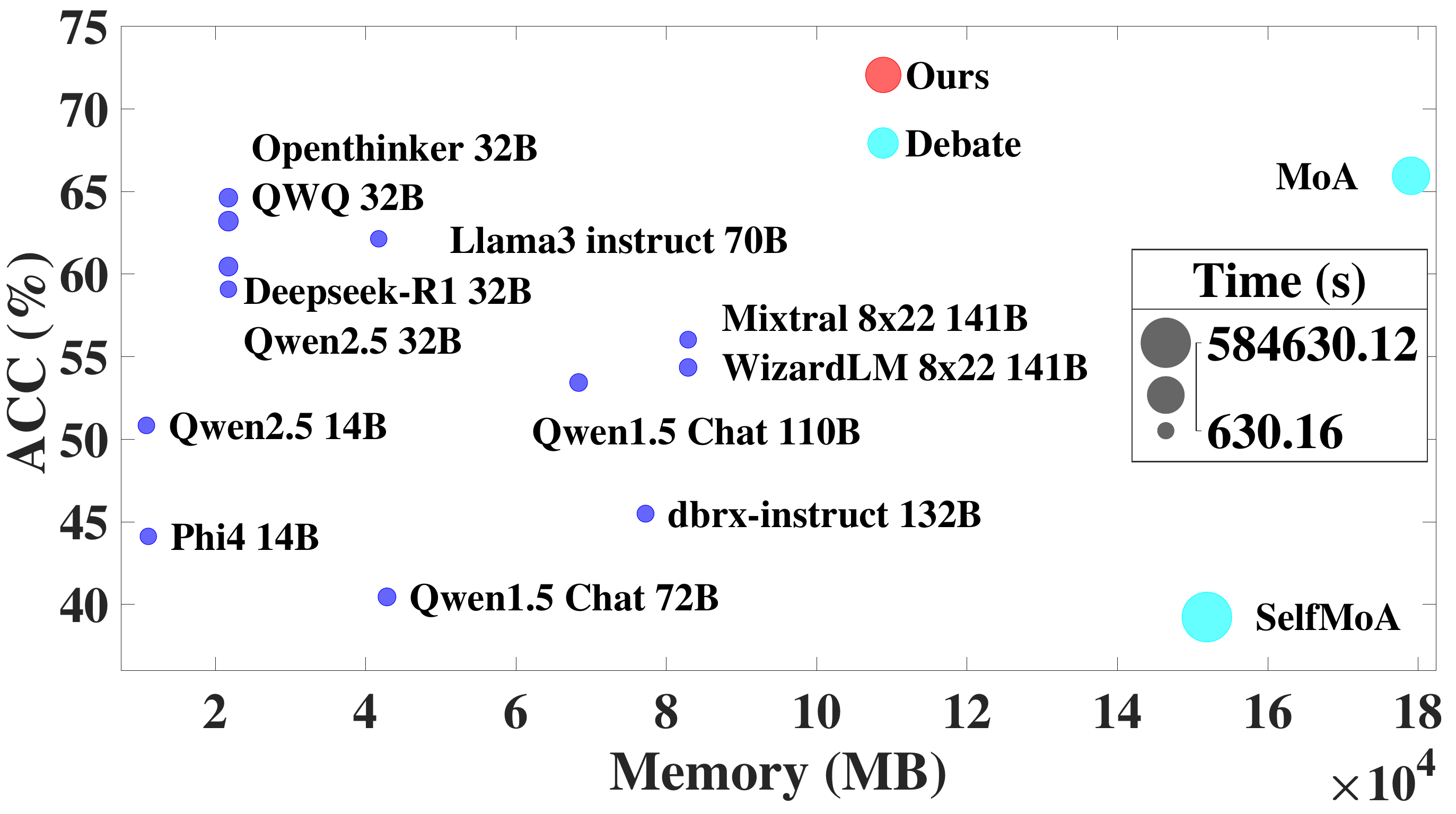}
    \caption{Comparisons of different LLMs and MAS-based models on diverse evaluation dimensions. Specifically, we evaluate the ACC (in percentage), occupied memory (in $10^4$ MB), and running time (in seconds) of different models on NEJMQA. The diameter of the bubble is proportional to the running time.} 
    \label{fig: motivation}
\end{figure}

Most recently, the emergence of large language models (LLMs) \cite{ouyang2022training,achiam2023gpt,chen2024more,liang2024can,yuksekgonul2025optimizing,wang2025towards} has substantially advanced artificial intelligence (AI), holding considerable promise for penetrating from general to domain-specific fields, with extreme interest in healthcare applications \cite{thirunavukarasu2023large,xu2023knowledge,li2024mediq,katz2024gpt,chen2025map}. 
Benefiting from the powerful representation ability of LLMs, the multi-agent system (MAS) \cite{li2024rethinking,kim2024mdagents,wang2025mixtureofagents,li2023camel,liang2024encouraging,chan2024chateval,zhang-etal-2024-exploring,estornell2024multi,feng2024don} has taken a tremendous step to effectively bridge the gap between AI development and healthcare deployment. Within this framework, multiple agents derived from LLMs generate a higher-quality output through reference and interaction.   
For example, Du \textit{et al.} \cite{du2023improving} pushed multiple model instances to propose, debate, and refine their reasoning responses over multiple rounds to enhance performance on complex tasks, reducing factual hallucinations \cite{sriramanan2024llm} that commonly exist in LLMs. Wang \textit{et al.} \cite{wang2025mixtureofagents} introduced a mixture-of-agents (MoA) framework, where a multi-layered ensemble of agents collectively enhances the response quality through an iterative process in which each agent utilizes all outputs from the preceding layer, enabling MoA to exceed GPT-4 \cite{achiam2023gpt}. Li \textit{et al.} \cite{li2024rethinking} developed an ensemble strategy that enhances inference performance by aggregating multiple outputs from a single top-performing agent through iterative rounds and further proposes a sequential version that dynamically aggregates outputs over multiple rounds. 

Nevertheless, MAS-based models supporting medical decision-making still present two key challenges due to their inherent limitations within agent construction and collaboration paradigm: 

{A major challenge in constructing an effective MAS-based model is the lack of a systematic pipeline for selecting an optimal set of collaborative agents from a vast and heterogeneous pool of candidate LLMs. Existing approaches often select agents solely based on standalone performance metrics, such as the accuracy \cite{wang2025mixtureofagents,li2024rethinking} or diversity \cite{chen2025self,wangtypedthinker}, while neglecting critical operational trade-offs essential for real-world clinical deployment. In practice, medical decision-making systems must balance competing objectives: high predictive performance, manageable computational resource consumption (often correlated with model size), and sufficient output information to foster robust decision-making. Indiscriminately employing multiple LLMs may yield marginal accuracy gains at unacceptable costs in inference time and throughput, hindering practical utility. Conversely, over-prioritizing cost by selecting only small, fast LLMs can compromise diagnostic reliability. Therefore, there is a pressing need for a mechanism that can rigorously navigate these multi-dimensional trade-offs and identify a Pareto-optimal cohort of agents, ensuring an optimal balance between system efficacy and operational efficiency in clinical environments.}

Another challenge is that current MAS-based models are constrained by their use of static collaboration patterns, lacking the flexibility necessary to adequately handle instances of collaboration failure. They are invalidated in medical decision-making scenarios, even yielding results inferior to individual LLMs. This phenomenon may stem from their candidate agents exhibiting overconfidence \cite{wen2024mitigating} in their incorrect outputs due to insufficient training caused by the scarcity of medical data, which propagates the generated misinformation to undermine their collaboration, leading to a significant performance decline. 
However, few studies have focused on the collaboration paradigm of the MAS-oriented framework in medical decision-making scenarios, yet improvements in their accessibility and accuracy could significantly reduce errors and optimize diagnosis pathways. Healthcare stands to benefit significantly from advances in the MAS-oriented framework, which complements rather than replaces physicians, particularly where experienced physicians are scarce \cite{keeler2006reducing,shen2021artificial,dvijotham2023enhancing,li2024mediq}.

To address these two aforementioned challenges, we propose a novel masked agent collaboration (MAC) framework that utilizes Pareto-optimal agent construction and cross-consistency maximization mechanisms to enhance the medical decision-making capacity, where the comparisons of single LLM and MAS-based framework architectures are given in Fig. \ref{fig: Comp_framework}. 
Specifically, we first perform a Pareto-frontier factor analysis on the pool of candidate LLMs by assessing model size, inference time, diversity score, and throughput ratio to achieve the fundamental trade-offs inherent towards real-world deployment. Within this analysis, we calculate the semantic similarity between pairwise outputs generated by each LLM to obtain its inherent diversity score, which facilitates the Pareto-optimal construction of agents. Such a multi-objective optimization guarantees that MAC is not only proficient but also practically effective, laying the groundwork for the subsequent dynamic MAS-oriented framework. 
Then, we calculate the output similarity between each pair of agents to determine their mutual consistency, quantitatively defined as the cross-consistency (CC) value, for the subsequent MAS-oriented cooperative stage. 
Afterward, we employ the CC maximization mechanism to implement a masking strategy that dynamically removes the agent with the minimum CC value, thereby discarding inconsistent or anomalous inputs for the next layer. 
Finally, we develop an adaptive progressive propagation architecture that enables the collective wisdom to evolve gradually, merging diverse perspectives while strengthening mutually consistent outcomes, where each operational agent in a specific layer integrates its output by consolidating and deliberating on the unmasked outputs from the prior layer.
This approach simulates a self-adjusting discussion, persistently enhancing the agreement of the group by segregating possibly untrustworthy viewpoints in every reasoning phase.

 The contributions of this work are summarized as follows:
\begin{itemize}
    \item We uncover the \textbf{weaknesses of current MAS-based frameworks} that they often fail in medical decision-making scenarios since underperforming expertise-agnostic LLMs easily introduce medical-agnostic misinformation towards collaboration, leading to unreliable and degraded diagnostic performance. 
    \item We develop the \textbf{Pareto-optimal agent construction mechanism} that evaluates candidate LLMs across multiple competing objectives, including model size, inference time, diversity score, and throughput, to systematically trade off performance and efficiency towards MAS-oriented adaptable real-world deployment.
    \item We propose the \textbf{cross-consistency maximization mechanism} to iteratively mask agents layer by layer, effectively mitigating performance degradation by filtering inconsistent outputs during the adaptive progressive propagation, thus resulting in reliable collaboration.
    \item We conduct \textbf{extensive experiments and evaluations} on three specialized medical datasets, demonstrating significant performance improvements on diverse evaluation dimensions while maintaining efficiency (Fig. \ref{fig: motivation}). 
\end{itemize}
    
\section{Related Work}
\subsection{LLMs in Reasoning}
\noindent {LLMs demonstrate a knowledge-integration and pattern-recognition capability in reasoning tasks, where they synthesize information from diverse training domains to generate contextually coherent and relevant outputs. Typically, they can be broadly categorized as general-domain and medical-domain ones, which are distinguished by their sources of training data and specialized knowledge base.}

\noindent\textbf{General-domain LLMs.} In the past decade, LLMs have increasingly performed complex domain-specific reasoning without extensive fine-tuning \cite{wei2022chain,zhou2022least,yao2023tree,besta2024graph}. This progress is primarily attributed to the emergence of reasoning techniques, which have become pivotal methods for enhancing the inferential capabilities of LLMs. 
For example, the chain-of-thought \cite{wei2022chain} addresses complex problems by guiding the model to generate a sequence of intermediate reasoning steps. Least-to-most \cite{zhou2022least} prompting decomposes a task into a series of subproblems solved in order, where the solution to each subproblem supports subsequent ones. Graph-of-thought \cite{besta2024graph} offers a more dynamic reasoning paradigm by modeling the reasoning process as a graph of interconnected thought nodes. Latest models, including Qwen3-VL \cite{yang2025qwen3}, Gemini2.5 \cite{comanici2025gemini}, and GPT-5 \cite{wang2025capabilities}, incorporate a human-like thinking mechanism, thereby allowing the depth of internal reasoning to be adaptively modulated according to the complexity of the task at hand.

\noindent\textbf{Medical-domain LLMs.} 
Recently, medical-domain LLMs, trained on large-scale medical training data and explicitly designed for clinical tasks, have similarly transitioned from a perception-alignment paradigm to one enhanced with explicit understanding and reasoning capabilities. Early models such as Med42 \cite{christophe2024med42} and HuatuoGPT-o1 \cite{zhang2023huatuogpt} have pioneered the extension of a general multimodal framework into the medical domain through targeted medical data adaptation and alignment. More recent domain-specific models, such as Medllama2 \cite{touvron2023llama}, have built upon this foundation by integrating medical semantic alignment and incorporating multi-step reasoning data during training, thereby strengthening reasoning and factual consistency in medical decision-making applications.

\subsection{Collaboration of MAS}

\noindent Recent studies have demonstrated that MAS-based frameworks can effectively integrate the respective strengths of LLMs, thereby achieving significant improvements in understanding and reasoning, especially in solving complex problems \cite{wang2022self,fu2023improving,chen2023reconcile,tang2024medagents,yang2023hierarchical,zhang2024proagent,li2024rethinking,kim2024mdagents,wang2024rethinking,wang2025mixtureofagents}. Within this framework, LLMs engage in a structured discussion, critique, or refinement over successive rounds, resulting in higher-performing solutions than those produced by a single LLM alone. This may be due to the fact that continuous exchange and synthesis of diverse perspectives foster deeper reasoning, mitigate individual model biases, and enhance the reliability and clinical applicability of AI-supported medical judgments. From an architectural perspective, existing MAS-based frameworks can be broadly categorized into two paradigms: role-playing and LLM-composition.

\noindent \textbf{Role‑playing paradigm}. Multiple LLMs are assigned distinct yet predefined roles, such as performer and supervisor, where each role focuses on a specific subtask aligned with its functional specialization \cite{kim2024mdagents,wang2024survey}. It allows for the systematic decomposition of complex medical problems and facilitates the integration of outputs from specialized models to enhance the overall coherence and comprehensiveness of the solution. For example, 
Li \textit{et al.} \cite{li2024agent} presented a simulacrum where LLM-based agents act as patients and doctors, enabling scalable, self-evolving clinical training through simulated treatment experiences. 
Tang \textit{et al.} \cite{tang2024medagents} proposed a dynamic expert evolution mechanism, yet their approach requires a large initial pool of experts and substantial computational resources for iterative refinement. 
Chen \textit{et al.} \cite{chen2025map} utilized a chief agent to coordinate triage, diagnosis, and treatment clinical agents to simulate and execute coherent inpatient care pathways.

\noindent \textbf{LLM-composition paradigm}. Research under this paradigm varies widely according to the strategy governing LLM selection, aggregation, and interaction.  Each LLM independently attempts to solve the given problem, after which their outputs are collectively analyzed to reach a consensus \cite{du2023improving}. 
For example, Du \textit{et al.} \cite{du2023improving} primarily focused on consensus‑building from independent outputs but overlooked the interplay between domain‑specific knowledge and cooperative dynamics among agents. 
Wang \textit{et al.} \cite{wang2025mixtureofagents} introduced a layered multi-agent framework where agents iteratively refine responses by integrating outputs from all agents in the preceding layer, achieving state-of-the-art results on multiple benchmarks. 
Bao \textit{et al.} \cite{bao2025expertise} introduced a two-stage process of expertise-aware agent recruitment and confidence-driven adversarial collaboration to enhance medical decision-making accuracy.

Nevertheless, there is an absence of a rigorous, principled methodology for selecting an optimal ensemble of collaborative agents from a vast, heterogeneous pool of candidate LLMs. Moreover, the reliance on predetermined, static collaboration patterns that lack the dynamic flexibility required to detect, mitigate, and adapt to instances of inter-agent failure or conflict.

\begin{figure*}[!]
    \centering
    \includegraphics [width=0.98 \linewidth]{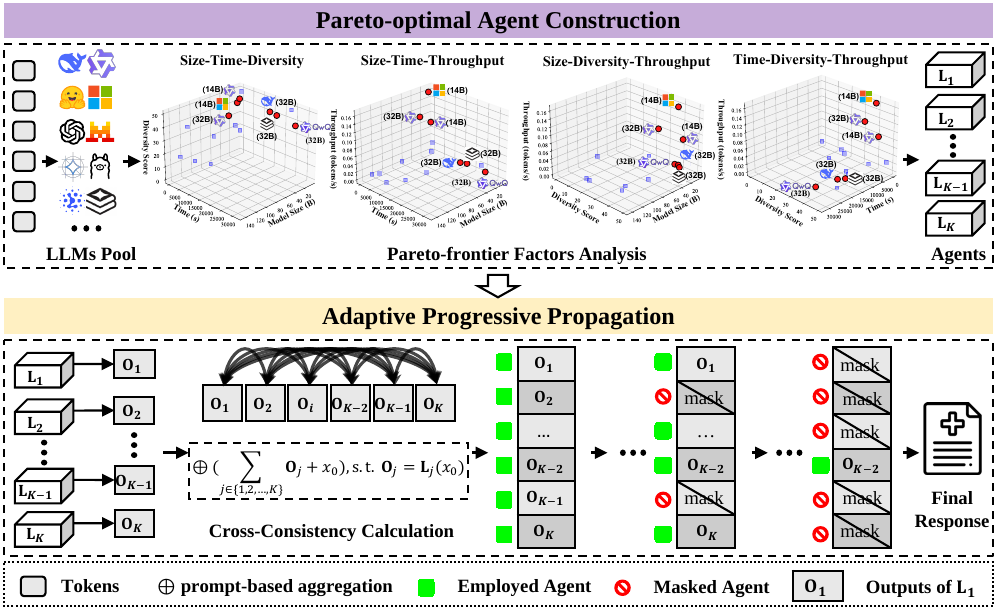}
    \caption{Illustration of the proposed MAC framework, where we achieve the Pareto-optimal agent construction via Pareto-frontier factors analysis and the adaptive progressive propagation via cross-consistency maximization mechanism. It significantly reduces the inconsistency of concatenated outputs while ensuring each LLM generates outputs based exclusively on outputs of unmasked LLMs from the previous layer as a contextual reference rather than considering entire outputs.}
    \label{fig: CC_framework}
  \end{figure*}

\section{Masked Agent Collaboration}

In this section, we introduce the proposed masked agent collaboration (MAC) framework, which achieves the Pareto-optimal agent construction via Pareto-frontier factors analysis and the adaptive progressive propagation via the cross-consistency maximization mechanism to enhance the medical decision-making capacity of MAC, as shown in Fig.~\ref{fig: CC_framework}.


 
\subsection{Pareto-optimal Agent Construction}
\begin{figure}
  \centering
  \includegraphics[width=0.92\linewidth]{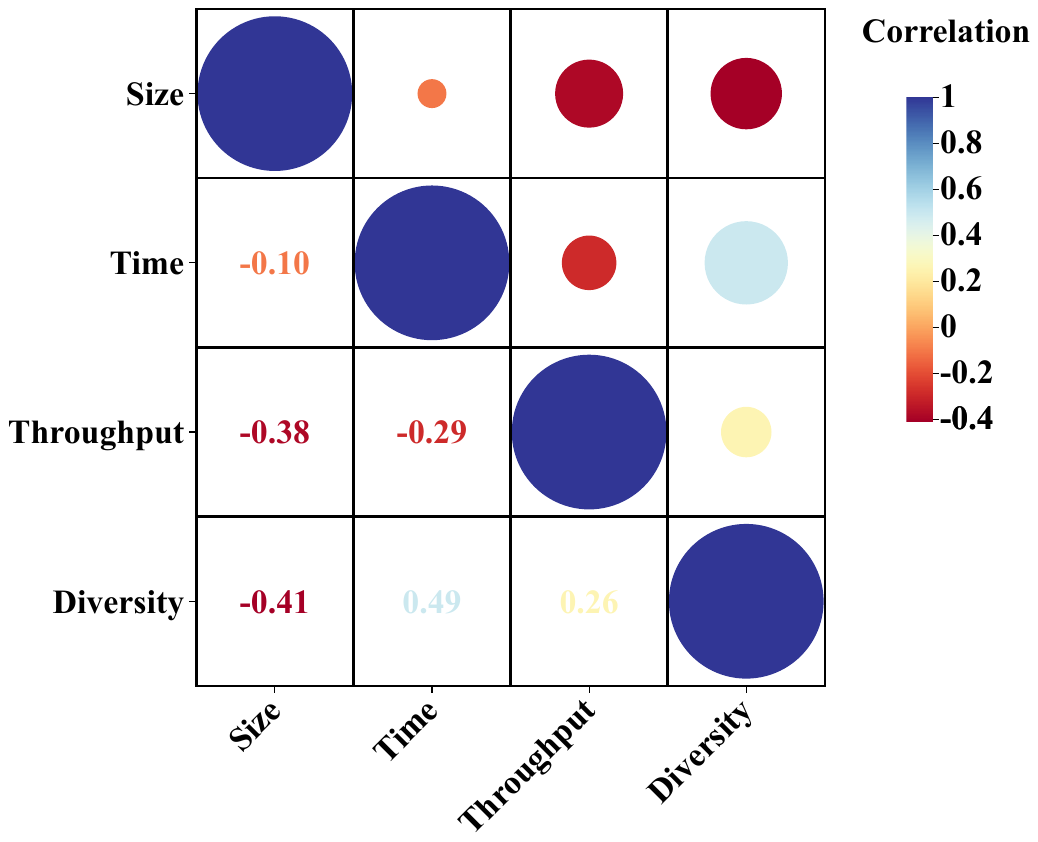}
  \caption{\textcolor{black}{The correlation matrix heatmap of the four key factors across the 15 candidate LLMs, where the absolute values of all correlation coefficients are well below the strong correlation threshold \cite{evans1996straightforward}.}}
  \label{fig:corr_heatmap}
\end{figure}

As aforementioned, existing MAS-based framework architectures \cite{kim2024mdagents,wang2025mixtureofagents,li2024rethinking} heavily rely on a predefined set of LLMs, which operates in the absence of a systematic framework for identifying an optimal ensemble of collaborative agents from an extensive and heterogeneous pool of candidate LLMs. In this work, we first consider four key factors of LLMs, including the model size, inference time, diversity score, and throughput ratio, to form a heuristic pipeline that captures the essential trade-offs in practical deployment. Specifically, model size and inference time represent critical resource constraints, i.e., storage and computational efficiency, while diversity score and throughput ratio represent task performance requirements, i.e., quality and creativity of outputs. Together, these factors define the Pareto-frontier where improving any one dimension inevitably compromises another, enabling systematic optimization of the efficiency-quality balance in real-world applications. \textcolor{black}{Fig. \ref{fig:corr_heatmap} is the correlation matrix heatmap of the four key factors across the 15 candidate LLMs, where the absolute values of all correlation coefficients are well below the strong correlation threshold (e.g., 0.6) \cite{evans1996straightforward}, indicating their rationality as separate factors in the optimization space.}

\subsubsection{Pareto-frontier Factors Analysis}
\label{subsec:Pareto}
Given a set of candidate LLMs $\mathcal{M} = \{ m_1, m_2, \dots, m_N \}$, each model $m_i$ is characterized by its model size $z_i$, inference time $t_i$, diversity score $d_i$ (details in Sec. \ref{subsec:diversity}), and token throughput $r_i = \tau_i/t_i$ with $\tau_i$ being its output token. Mathematically, we formulate the Pareto-optimal selection problem as a multi-objective optimization aiming to simultaneously minimize $z_i$ and $t_i$ while maximizing $d_i$ and $r_i$, where the model $m_i$ is said to dominate $m_j$ (denoted $m_i \prec m_j$) if it is superior in at least one objective and not inferior in any other. That is, $(z_i \leq z_j) \land (t_i \leq t_j) \land (d_i \geq d_j) \land (r_i \geq r_j) \land \left[ (z_i < z_j) \lor (t_i < t_j) \lor (d_i > d_j) \lor (r_i > r_j) \right]$. Specifically, the Pareto frontier is defined as:
\begin{equation}
    \begin{aligned}
        F_1 &= \left\{ m_i \in \mathcal{M} \mid \nexists  m_j \in \mathcal{M} : m_j \prec m_i \right\}, \\
        F_k &= \left\{ m_i \in \mathcal{M} \setminus \bigcup_{\ell=1}^{k-1} F_\ell \;\middle|\; \right. \\
        &\qquad \left. \nexists m_j \in \mathcal{M} \setminus \bigcup_{\ell=1}^{k-1} F_\ell : m_j \prec m_i \right\}, \quad k > 1.
    \end{aligned}
    \label{eq:def_F}
\end{equation}

To construct a diverse and complementary ensemble of $K$ models, we employ a coverage-maximization strategy in the normalized feature space. Each model is represented as a vector $\mathbf{x}_i$ where all objectives are converted to minimization and normalized to $[0,1]$, as follows:
\begin{equation}
\mathbf{x}_i = \left[ \frac{f_k(m_i) - \min f_k}{\max f_k - \min f_k} \right], \quad \rm{s.t.} \quad {\mathit{k} \in \mathcal{K}},
\label{eq:feature_vector}
\end{equation}
where $\mathcal{K} = \{z, t,  1-d, 1-r\}$, $f_z = z$, $f_t = t$, $f_{1-d} = -d$, $f_{1-r} = -r$. 
More specifically, the selection process begins by constructing the Pareto-optimal front $F_1$ through a systematic pairwise dominance comparison among all models $\mathcal{M}$. For each model $m_i$, we verify whether there exists any other model $m_j$ that dominates it, i.e., $m_j$ is at least as good as $m_i$ on all metrics and strictly better on at least one metric. It is formally expressed as $\forall k, \mathbf{x}_j[k] \geq \mathbf{x}_i[k]$ and $\exists k, \mathbf{x}_j[k] > \mathbf{x}_i[k]$. Models not dominated by any other are included in $F_1$, representing the set of non-dominated candidates in the multi-objective space. In this work, the models in $F_1$ are sorted in descending order of throughput $r_i$ to prioritize high-efficiency candidate LLMs since there is a notable degradation in their reasoning performance at much shorter input lengths than their technical maximum \cite{levy2024same}. We then proceed with a two-phase selection strategy, where up to $K$ models (or fewer if $|F_1| < K$) are selected directly from the Pareto front $F_1$, and these constitute the initial ensemble $\mathcal{S}$. If the size of $\mathcal{S}$ remains below $K$, we supplement it by selecting the remaining highest-throughput models from the remaining pool $\mathcal{M}_{\text{remain}} = \mathcal{M} \setminus \mathcal{S}$, again ordered by $r_i$. It can be interpreted as starting with the Pareto-optimal set $F_1$ and iteratively adding the model that maximizes the incremental coverage gain until $K$ models are selected. The approach is theoretically justified by the sub-modularity of the coverage function, which ensures that the algorithm achieves the optimal coverage. In practice, applied to the experimental dataset, the algorithm first identifies the initial Pareto front $F_1$, comprising several top-performing models. Subsequent iterations then select additional models that maximally expand the coverage in the time-diversity trade-off regions, yielding the final ensemble $\mathcal{S}$. 

Overall, this methodology ensures a principled, objective selection of complementary models without reliance on ad hoc weighting schemes, making it particularly suitable for multi-faceted evaluation scenarios where balanced performance across efficiency, quality, and diversity is desired. Notably, in practice, the coverage of the selected model ensemble $\mathcal{S}$ can be quantified as the weighted sum of ranges across the four dimensions, i.e., $ \sum_{\mathit{k} \in \mathcal{K}} w_k \cdot \left[  \frac{f_k(m_i) - \min f_k}{\max f_k - \min f_k} \right]$, where $w_k$ represents the weight assigned to the dimension $k$. It effectively measures the span of the selected LLMs in the multi-objective optimization space, capturing their collective ability to represent diverse regions of the performance trade-off surface. In this work, equal weights $w = [1.0, 1.0, 1.0, 1.0]^\top$ are applied, indicating balanced consideration across all dimensions. 


\subsubsection{Diversity Quantitative Calculation}
\label{subsec:diversity}
We calculate the diversity score of LLMs by computing the similarity between their pairwise outputs. For example, for the individual LLM $\mathbf{L}_{\bf{1}}$, we sample ten outputs from it to one query, denoted as $\{\mathbf{O}_{\bf{1}}^{j}\}_{j=1}^{10}$. For any given pair of outputs, take $\mathbf{O}_{\bf{1}}^{1}$ and $\mathbf{O}_{\bf{1}}^{2}$ ($|\mathbf{O}_{\bf{1}}^{1}| \leq |\mathbf{O}_{\bf{1}}^{2}|$) as an example, we compute their similarity by finding the best matching substring of $ \mathbf{O}_{\bf{1}}^{2} $ that aligns with $ \mathbf{O}_{\bf{1}}^{1} $. For each position $ i \in \mathbf{O}_{\bf{1}}^{2} $, the substring is obtained as follows:
\begin{equation} 
\begin{aligned}
\label{eq:1}
& \mathbf{O}_{\bf{1}}^{sub} = \mathbf{O}_{\bf{1}}^{2}[i:i + |\mathbf{O}_{\bf{1}}^{1}|], \\
& \mathrm{s.t.} \quad i \in \{0, \dots, |\mathbf{O}_{\bf{1}}^{2}| - |\mathbf{O}_{\bf{1}}^{1}|\},
\end{aligned}
\end{equation}
where $ \mathbf{O}_{\bf{1}}^{1} $ slides over $ \mathbf{O}_{\bf{1}}^{2} $ with a window of size $ |\mathbf{O}_{\bf{1}}^{1}| $. The similarity value of each window can be computed as follows
\begin{equation}
\label{eq:2}
  \underset{{s\in\mathbf{O}_{\bf{1}}^{sub}}}{sim}(\mathbf{O}_{\bf{1}}^{1}, s) = \left(1 - \frac{D(\mathbf{O}_{\bf{1}}^{1}, s)}{|\mathbf{O}_{\bf{1}}^{1}|}\right) \times 100,
\end{equation}
where $ D(\mathbf{O}_{\bf{1}}^{1}, \mathbf{O}_{\bf{1}}^{sub}) $ denotes the Levenshtein distance of $\mathbf{O}_{\bf{1}}^{1}$ and $\mathbf{O}_{\bf{1}}^{sub}$. We use that distance metric since it is useful for detecting partial matches in string data  \cite{levenshtein1966binary}. Afterward, the output diversity of $\mathbf{O}_{\bf{1}}^{1}$ and $\mathbf{O}_{\bf{1}}^{2}$ can be computed as:
\begin{equation}
\label{eq:3}
{div}(\mathbf{O}_{\bf{1}}^{1}, \mathbf{O}_{\bf{1}}^{2}) = 100 - \max \left( \underset{{s\in\mathbf{O}_{\bf{1}}^{sub}}}{sim}(\mathbf{O}_{\bf{1}}^{1}, s) \right).
\end{equation}
Similarly, we can obtain their values for all other pairwise outputs in $\{\mathbf{O}_{\bf{1}}^{j}\}_{j=1}^{10}$, resulting in a total number for the permutation of $\{\mathbf{O}_{\bf{1}}^{j}\}_{j=1}^{10}$ (that is, $C_{10}^{2}$). We take the mean of the above values as the diversity score for $\mathbf{L}_{\bf{1}}$, where the diversity score of the LLM is higher, its output is more diverse. In this way, we can trade off the diversity score of candidate LLMs within Pareto-frontier factors analysis to achieve agent construction for the subsequent masked agent collaboration.

\begin{algorithm}[!]
    \caption{Adaptive Progressive Propagation}
    \label{alg: methodology}
    \begin{algorithmic}[1]
        \REQUIRE 
            Input data $\mathbf{x}_0$,  pre-selected candidate models $\{\mathbf{L}_{\bf{1}}, \mathbf{L}_{\bf{2}}, \mathbf{L}_{\bf{3}}, \mathbf{L}_{\bf{4}}, \mathbf{L}_{\bf{5}}, \mathbf{L}_{\bf{6}}\}$, number of collaboration layers $l$, iteration counter initialization $i \gets 1$
        \ENSURE Final result $\mathbf{R}_l$
        \STATE Set initial representation $\mathbf{R}_{0} \gets \mathbf{x}_0$
        \STATE Initialize active agent indices $\mathbf{L}_{idx}$ 
        \STATE Initialize output storage $\mathbf{O} \gets \varnothing$
        \FORALL{$j \in \mathbf{L}_{idx}$}
            \STATE Compute agent output $\mathbf{O}_{j}^{1} \gets \mathbf{L}_{j}(\mathbf{R}_{0})$ 
            \STATE Store output $\mathbf{O}[j] \gets \mathbf{O}_{j}^{1}$
        \ENDFOR
        \STATE Update $\mathbf{R}_1$ via Eq.~(\ref{eq: out_1stlayer})
        \STATE For all $j \in \mathbf{L}_{idx}$, let $\mathbf{O}_{j} \equiv \mathbf{O}_{j}^{1}$ 
        \WHILE{$i < l$}
            \STATE Initialize similarity matrix $\mathbf{S} \gets \mathbf{0}_{|\mathbf{L}_{idx}| \times |\mathbf{L}_{idx}|}$
            \FORALL{distinct pairs $(p,q) $, where $p \neq q$}
                \STATE Update $\mathbf{S}[p,q]$ via Eq.~(\ref{eq:3})
            \ENDFOR
            \STATE Update $\mathbf{c}$ via Eq.~(\ref{eq:cc_min})
            \STATE $\mathbf{L}_{idx} \gets \mathbf{L}_{idx} \setminus \{\mathbf{c}\}$
            \STATE $\mathbf{O} \gets \mathbf{O} \setminus \{\mathbf{O}_{\mathbf{c}}\}$
            \STATE Update agent outputs with current representation
            \FORALL{$j \in \mathbf{L}_{idx}$}
                \STATE $\mathbf{O}_{j} \gets \mathbf{L}_{j}(\mathbf{R}_i)$
            \ENDFOR
            \STATE Compute next layer representation
            \STATE Update $\mathbf{R}_{i+1} $ via Eq.~(\ref{eq: out_layer})
            \STATE Update iteration counter $i \gets i + 1$
            \STATE Update representation $\mathbf{R}_i \gets \mathbf{R}_{i+1}$
        \ENDWHILE
        \STATE \textbf{return} final collaboratively refined output $\mathbf{R}_l$ 
    \end{algorithmic}
\end{algorithm}

\subsection{Adaptive Progressive Propagation}

Previous methods \cite{wang2025mixtureofagents, li2024rethinking} achieve the collaboration among agents through the iterative aggregation of all outputs, where each layer aggregates the complete outputs of every agent from the previous layer to propagate information. This approach inevitably introduces interference from low-quality, redundant outputs and incurs substantial computational time, which severely constrains the scalability and practicality of MAS-based frameworks, particularly in real-time applications or those involving lengthy, context-dependent generation tasks. The fundamental issue lies in the indiscriminate treatment of all agent contributions, allowing outputs with high variance, logical inconsistencies, or factual inaccuracies to be propagated and amplified through the collaboration chain. This not only degrades the final output quality but also squanders computational resources on processing and integrating information from unreliable sources.

To address this issue, we propose an adaptive progressive propagation via a cross-consistency maximization mechanism that iteratively excludes the least consistent agent layer by layer, thereby refining the collaborative process. The implementation consists of two key steps, including measuring pairwise cross-consistency (CC) values between agents and masking the agent with the lowest pairwise CC score to propagate the outputs of the remaining agents. In this way, our method introduces a dynamic, data-driven filtering process that actively identifies and mitigates the influence of outliers during the collaborative inference, rather than relying on a static, fixed architecture. This selective propagation ensures that the information flow is progressively purified, focusing computational effort on the most coherent and mutually reinforcing agent outputs. The core hypothesis is that agents producing outputs that are highly divergent from the consensus are likely introducing noise or errors; systematically pruning them leads to a more robust and efficient convergence towards high-quality solutions.

Specifically, in each layer, every agent generates its output based on the aggregated outputs of the unmasked agents from the previous layer. This creates a self-reinforcing loop where higher-consistency cohorts iteratively refine their collaborative reasoning. The framework is formalized mathematically as follows. Let \(\mathbf{L}_{1}, \mathbf{L}_{2}, \mathbf{L}_{3}, \mathbf{L}_{4}, \mathbf{L}_{5}, \mathbf{L}_{6}\) denote the pre-selected agents, the first-layer representation \(\mathbf{R}_1\) is obtained by:
\begin{equation}
\mathbf{R}_1 = \bigoplus \biggl( \sum_{j \in \mathbf{L}_{idx}} \mathbf{O}_{j}^{1} + \mathbf{\textit{x}}_0 \biggr),
\quad
\text{s.t.} \ 
\biggl\{
\begin{aligned}
&\mathbf{R}_{0} = \mathbf{\textit{x}}_0, \\
&\mathbf{O}_{j}^{1} = \mathbf{L}_{j}(\mathbf{R}_{0}),
\end{aligned}
\biggr.
\label{eq: out_1stlayer}
\end{equation}
where \(\mathbf{L}_{idx} = \{ 1, 2, \ldots, K-1, K \}\) is the set of agent indices, \(\mathbf{x}_0\) is the input, \(+\) denotes the output concatenation, \(\sum\) denotes the output concatenation, \(\mathbf{L}_{j}(\mathbf{R}_{0})\) is the output of the \(j\)-th agent given input \(\mathbf{R}_{0}\), and \(\bigoplus(\cdot)\) represents prompt-based aggregation. For readability, we denote the first-layer outputs \(\mathbf{O}_{1}^{1}, \ldots, \mathbf{O}_{K}^{1}\) simply as \(\mathbf{O}_{1}, \ldots, \mathbf{O}_{K}\) in subsequent steps.

After obtaining the initial outputs, we compute the pairwise CC values between each pair of distinct agents using a similarity function \(sim(\cdot,\cdot)\) defined in Eq. (\ref{eq:3}). This function quantitatively assesses the semantic or logical alignment between two agent outputs, serving as the basis for identifying dissonance within the group. The agent with the lowest minimum pairwise similarity is identified as the least consistent agent:
\begin{equation}
\mathbf{c} = \mathop{\arg\min}_{k \in \mathbf{L}_{idx}} \left( \min_{j \in \mathbf{L}_{idx}, j \neq k} \mathbf{S}[k,j] \right),
\label{eq:cc_min}
\end{equation}
where \(\mathbf{S}[k,j]\) denotes the similarity value of $\mathbf{O}_{k}$ and $\mathbf{O}_{j}$. This agent is then masked to remove it from the active set \(\mathbf{L}_{idx}\), where its output is excluded from further aggregation. This targeted removal acts as a form of iterative refinement, where the collaborative system becomes increasingly homogeneous and aligned with each pruning step, effectively reducing the solution space to regions of higher agent agreement and presumed higher quality.

\begin{figure}
  \centering
  \includegraphics[width=0.98\linewidth, height = 1.68\linewidth]{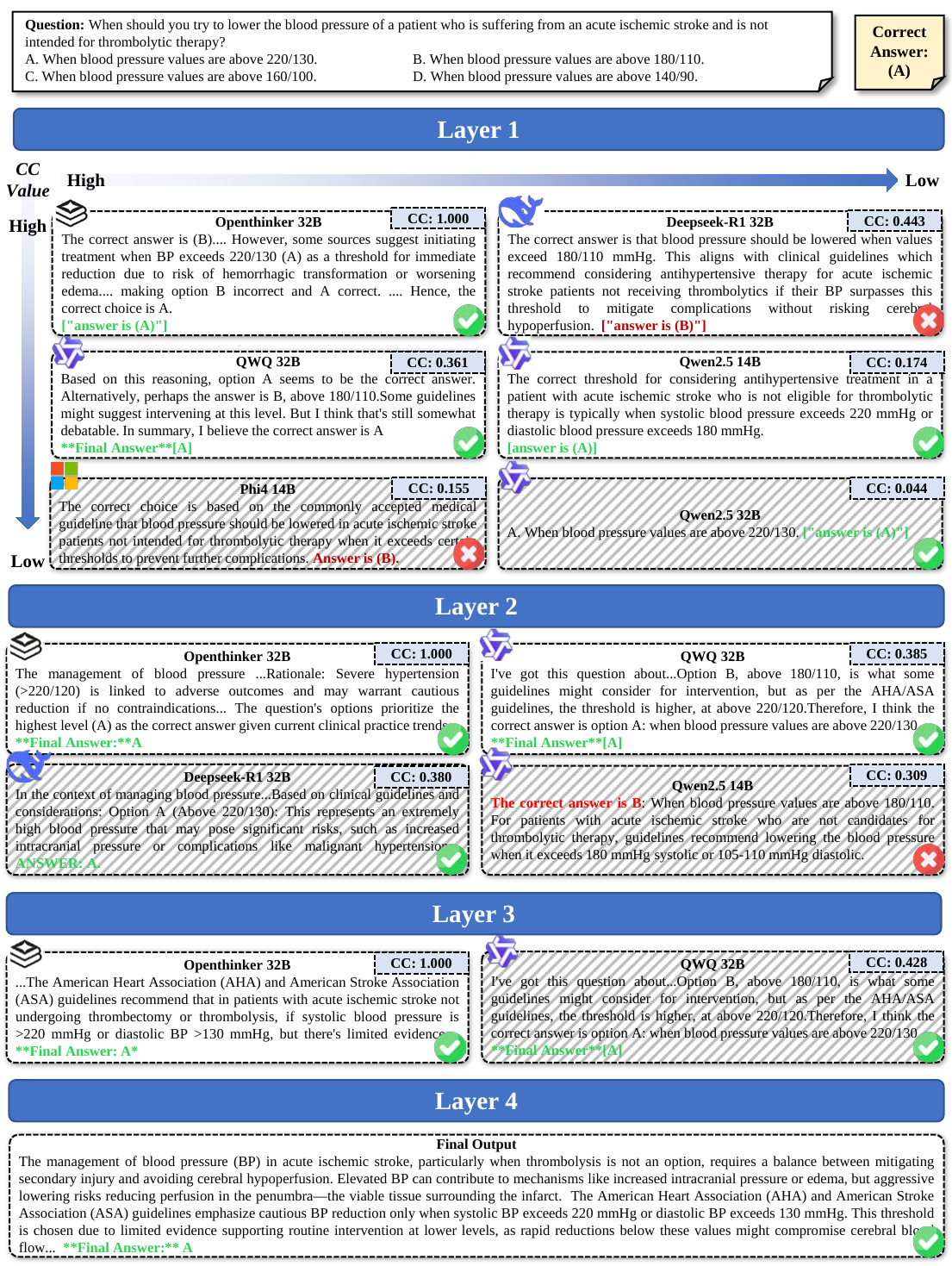}
  \caption{An overview of the adaptive progressive propagation of MAC, where MAC fosters collaboration through the adaptive exclusion of inconsistent outputs to augment medical decision-making capabilities. The green font highlights correct details, while the red font indicates hallucinations or inaccuracies in LLM responses, and the area covered by the diagonal dashed line denotes that the agent is being masked.}
  \label{fig:case}
\end{figure}

In the subsequent layer, the remaining agents generate new outputs conditioned on the current aggregated representation \(\mathbf{R}_i\). The updated representation for layer \(i\) is computed as:
\begin{equation}
\mathbf{R}_i = \bigoplus \biggl( \sum_{j \in \mathbf{L}_{idx} \setminus \{\mathbf{c}\}} \mathbf{L}_{j}(\mathbf{R}_{i-1}) + \mathbf{R}_{i-1} \biggr),
\label{eq: out_layer}
\end{equation}
where \(\sum_{j \in \mathbf{L}_{idx} \setminus \{\mathbf{c}\}}\) denotes the concatenation of outputs from all active agents except the masked one. This process repeats iteratively, where we recompute pairwise similarities among the currently active agents at each layer, mask the one with the lowest consistency, and propagate the outputs of the remaining agents. The collaboration continues until a predefined number of layers \(l\) is reached, yielding the final refined output \(\mathbf{R}_l\). This design allows the collaboration depth \(l\) to be a flexible hyperparameter, offering a direct trade-off between computational cost (fewer active agents per layer) and the breadth of considered perspectives (more agents retained). The complete procedure, summarized in Algorithm \ref{alg: methodology}, enables dynamic pruning of inconsistent agents throughout the collaboration, reducing noise from low-quality outputs and improving both efficiency and robustness in multi-agent inference. 


\textcolor{black}{We provide a comprehensive case study of our MAC on NEJMQA to qualitatively demonstrate the entire workflow of the adaptive progressive propagation across five core stages, as shown in Fig. \ref{fig:case}. Specifically, the process begins with \textit{agent selection}, initializing a pool of diverse agents selected via our PAC mechanism (e.g., Openthinker 32B, Deepseek-R1 32B, QWQ 32B, Qwen2.5 32B, Qwen2.5 14B, and Phi4 14B). In each layer, the framework generates \textit{intermediate outputs}, where each active agent follows a JavaScript Object Notation (JSON) \cite{bray2014javascript} prompt template to articulate its reasoning process and decision. Subsequently, we calculate the \textit{cross-consistency scores} to quantitatively measure the semantic alignment among agents (e.g., CC: 0.443 for Deepseek-R1 32B). Based on these scores, the framework executes \textit{masking decisions}; by sorting CC values in descending order, the agent with the lowest CC score is rigorously removed at each iteration (visually indicated by diagonal dashed lines). The remaining agents then proceed to conduct collaboration in the next layer. This iterative refinement culminates in the \textit{final diagnosis}, successfully correcting initial hallucinations to reach an accurate medical conclusion. By adaptively eliminating inconsistent outputs to ensure robust collaboration, this case study demonstrates that our MAC framework effectively enhances reliable medical decision-making, outperforming individual LLMs and other MAS-based models.}

\subsection{Time Complexity}
The time complexity of Algorithm~\ref{alg: methodology} is given by $O\big(l K^2 T_{\text{sim}} + l  K  T_{\text{infer}}\big)$, where $l$ denotes the number of collaboration layers, $K$ is the initial number of candidate agents, and $T_{\text{sim}}$ represents the cost of computing pairwise similarity between two outputs. As defined in Eq.~(\ref{eq:3}), it involves sliding‑window substring matching and Levenshtein distance computation with worst‑case complexity $O(L^3)$ for outputs of length $L$. $T_{\text{infer}}$ is the inference time of a single agent. In each iteration, the algorithm performs $O(K^2)$ pairwise similarity calculations, identifies and masks the least consistent agent, 
and updates the outputs of the remaining agents through inference. Because $K$ decreases after each masking step, the total cost over $l$ layers is dominated by the similarity computations, resulting in an overall polynomial complexity of $O\big(l K^2 L^3 + l K T_{\text{infer}}\big)$. 


\section{Experiments}


\begin{table}[!]
\centering
\caption{Question length and option length are calculated in tokens. “Avg./Max. que./opt.” represents “Average/Maximum question/option length”.}
\resizebox{0.99\columnwidth}{!}{
\resizebox{1\columnwidth}{!}{
\begin{tabular}{cccccc}
\hline\hline
Dataset & Samples & Options  & Avg./Max. que. & Avg./Max. opt. \\
\hline
NEJMQA  & 655     & 4                 & 33.46 / 163           & 5.01 / 31             \\
MMLUPH  & 818     & 10                & 43.81 / 728           & 5.71 / 28             \\
MedQA   & 1273    & 5                & 142.07 / 551          & 3.49 / 36             \\
\hline\hline
\end{tabular}
}
}
\label{tab: datasets}
\end{table}

\subsection{Experimental Setting}
\noindent \textbf{Datasets.} To evaluate the medical decision-making capacity of LLMs, we employ three publicly available medical datasets:  NEJMQA \cite{katz2024gpt}, MMLUPH \cite{wang2024mmlu}, and MedQA \cite{jin2021disease}, as summarized in Table \ref{tab: datasets}. 

\noindent \textbf{NEJMQA} is derived from Israel's 2022 medical specialist licensing examination, covering five core clinical disciplines: \textit{General Surgery, Internal Medicine, Obstetrics and Gynecology, Pediatrics,} and \textit{Psychiatry}. Notably, physicians must achieve a minimum passing score of 65\% in each discipline to obtain board certification. 

\noindent \textbf{MMLUPH}, a health topic of the MMLU-Pro, contains 818 carefully curated questions that span eight medical specialties: \textit{Virology, Professional Medicine, Nutrition, Medical Genetics, Human Aging, College Medicine, Clinical Knowledge,} and \textit{Anatomy}. 

\noindent\textbf{MedQA} is a specialized question-answering benchmark designed to evaluate and advance the capabilities of LLMs, which consists of a large collection of queries sourced from real-world medical examinations, as well as queries reflecting clinical practice. It covers a broad spectrum of medical topics, such as \textit{diagnosis, treatment, and pharmacological information}. The 1273 queries and answers from the test set are used for evaluation.

\noindent \textbf{Models.} 
{To gain a deeper understanding of the performance advantages of our method, we conduct comparisons with fifteen open-access models (Medllama2 7B \cite{touvron2023llama}, Med42 8B \cite{christophe2024med42}, HuatuoGPT-o1 8B \cite{zhang2023huatuogpt}, Phi4 14B \cite{abdin2024phi}, Qwen2.5 14B, Qwen2.5 32B \cite{qwen2.5}, QWQ 32B \cite{qwq32b}, Openthinker 32B \cite{openthoughts}, Deepseek-R1 32B \cite{guo2025deepseek}, Llama3 instruct 70B \cite{meta2024introducing}, Qwen1.5 Chat 72B, Qwen1.5 Chat 110B \cite{bai2023qwen}, dbrx-instruct 132B \cite{team2024introducing}, Mixtral 8x22 141B \cite{jiang2024mixtral}, and WizardLM 8x22 141B \cite{xu2023wizardlm}), three closed-source models (Claude Haiku 4.5 \cite{anthropic2025haiku}, GPT-5.1 \cite{openai2025gpt5systemcard}, and Gemini 2.5 Flash \cite{comanici2025gemini25}), and three MAS-based models (Debate \cite{du2023improving}, SelfMoA \cite{li2024rethinking}, and MoA \cite{wang2025mixtureofagents}). We also developed a variant called Ours w/ GPTs, which prioritizes high-quality outputs by using heterogeneous LLMs Claude Haiku 4.5, GPT-5.1, and Gemini 2.5 Flash as the collaborative agents to evaluate the effectiveness of the proposed adaptive progressive propagation architecture.}

\noindent \textbf{Implementation Details.} 
Existing MAS-based models \cite{du2023improving,wang2025mixtureofagents,li2024rethinking} heavily rely on a predefined set of LLMs, where the size of an individual LLM can reach 141B parameters, imposing severe limitations on clinical deployment due to high computational demands. 
Based on the proposed Pareto-optimal agent construction, our framework exclusively employs open-access LLMs ranging from 14B to 32B parameters, where the maximum 32B parameter LLM requires 21,735 MB of GPU memory, equivalent to one NVIDIA GeForce RTX 4090, making the configuration both practical and cost-effective. The specific agents for the collaboration of LLMs include Phi4 14B, Qwen2.5 14B, Qwen2.5 32B, QWQ 32B, Openthinker 32B, and Deepseek-R1 32B in our framework. 


\begin{table*}[!]
\centering
\caption{Evaluations with five evaluation metrics on NEJMQA, demonstrating substantial performance improvements with our method in medical decision-making scenarios. `Ours w/ GPTs' corresponds to using Claude Haiku 4.5, GPT-5.1, and Gemini 2.5 Flash as the collaborative agents in MAC. We highlighted the best results with \textbf{bold}, the second-best open-access results with \underline{underline}.}
\resizebox{1.98\columnwidth}{!}{
\begin{tabular}{l|l|lllll}
\hline\hline
Type                          & LLMs                & ACC         & F1          & PRE         & SPE         & MCC         \\
\hline
\textcolor{black}{\multirow{4}{*}{Closed-source} }& \textcolor{black}{ Claude Haiku 4.5                   }& \textcolor{black}{ 64.58$\pm$1.06 }& \textcolor{black}{ 64.75$\pm$1.02 }& \textcolor{black}{ 65.07$\pm$0.96 }& \textcolor{black}{ 92.34$\pm$1.79 }& \textcolor{black}{ 52.73$\pm$1.38} \\
& \textcolor{black}{ GPT-5.1                            }& \textcolor{black}{ 76.28$\pm$1.47 }& \textcolor{black}{ 76.28$\pm$1.48 }& \textcolor{black}{ 76.54$\pm$1.43 }& \textcolor{black}{ 92.09$\pm$0.48 }& \textcolor{black}{ 68.37$\pm$1.93} \\
& \textcolor{black}{ Gemini 2.5 Flash                   }& \textcolor{black}{ \underline{78.57$\pm$0.73} }& \textcolor{black}{ \underline{78.69$\pm$0.62} }& \textcolor{black}{ \underline{78.88$\pm$0.50} }& \textcolor{black}{ \underline{93.60$\pm$0.57} }& \textcolor{black}{ \underline{71.36$\pm$0.94}} \\
& \textcolor{black}{ Ours w/ GPTs }& \textcolor{black}{ \textbf{80.65$\pm$1.05} }& \textcolor{black}{ \textbf{80.77$\pm$1.10} }& \textcolor{black}{ \textbf{81.01$\pm$1.19} }& \textcolor{black}{ \textbf{94.39$\pm$1.03} }& \textcolor{black}{ \textbf{74.14$\pm$1.45} }\\
\hline
\multirow{19}{*}{Open-access} & Medllama2 7B        & 38.47$\pm${3.16}    & 37.76$\pm${3.59}    & 39.25$\pm${2.88}    & 84.31$\pm${5.18}    & 18.42$\pm${2.51}    \\
& Med42 8B            & 46.67$\pm${1.60}    & 46.53$\pm${2.01}    & 47.03$\pm${2.39}    & 87.75$\pm${2.98}    & 29.83$\pm${1.17}    \\
& HuatuoGPT-o1 8B     & 30.08$\pm${17.32}   & 33.26$\pm${13.33}   & 42.99$\pm${4.33}    & 85.07$\pm${5.91}    & 17.53$\pm${15.28}   \\
& Phi4 14B            & 55.88$\pm${10.18}   & 55.91$\pm${10.28}   & 59.44$\pm${4.92}    & 86.68$\pm${1.07}    & 42.55$\pm${11.23}   \\
& Qwen2.5 14B         & 54.71$\pm${3.52}    & 54.95$\pm${3.21}    & 55.44$\pm${2.88}    & 87.01$\pm${1.14}    & 39.67$\pm${4.50}    \\
& Qwen2.5 32B         & 61.47$\pm${2.11}    & 61.43$\pm${2.11}    & 61.71$\pm${1.81}    & 87.09$\pm${0.65}    & 48.45$\pm${2.64}    \\
& QWQ 32B             & 65.24$\pm${1.79}    & \underline{65.24$\pm${1.81}}   & 65.62$\pm${2.15}    & 89.13$\pm${1.77}    & 53.58$\pm${2.55}    \\
& Openthinker 32B     & 62.65$\pm${2.02}    & 62.71$\pm${2.02}    & 63.48$\pm${2.12}    & 88.46$\pm${0.94}    & 50.33$\pm${2.65}    \\
& Deepseek-R1 32B     & 61.63$\pm${1.04}    & 61.69$\pm${1.04}    & 62.09$\pm${0.91}    & 88.92$\pm${1.42}    & 48.80$\pm${1.38}    \\
& Llama3 instruct 70B & 60.97$\pm${1.04}    & 60.99$\pm${1.01}    & 61.41$\pm${1.02}    & 88.75$\pm${1.19}    & 47.87$\pm${1.39}    \\
& Qwen1.5 Chat 72B    & 41.17$\pm${1.51}    & 41.57$\pm${1.57}    & 43.10$\pm${1.66}    & 84.41$\pm${0.42}    & 21.99$\pm${2.06}    \\
& Qwen1.5 Chat 110B   & 53.89$\pm${0.79}    & 54.21$\pm${0.93}    & 55.09$\pm${1.11}    & 87.74$\pm${0.25}    & 38.58$\pm${1.19}    \\
& dbrx-instruct 132B  & 47.69$\pm${2.80}    & 46.91$\pm${2.82}    & 48.38$\pm${2.35}    & 85.95$\pm${0.67}    & 30.09$\pm${3.42}    \\
& Mixtral 8x22 141B   & 55.93$\pm${0.32}    & 55.98$\pm${0.38}    & 56.38$\pm${0.47}    & 87.27$\pm${1.80}    & 41.19$\pm${0.46}    \\
& WizardLM 8x22 141B  & 56.39$\pm${2.54}    & 57.04$\pm${2.17}    & 58.44$\pm${1.87}    & 86.64$\pm${1.43}    & 42.34$\pm${3.11}    \\
& Debate [ICML'23]    & \underline{66.41$\pm${1.61}} & 64.86$\pm${2.66} & \underline{65.84$\pm${3.27}} & \underline{90.97$\pm${1.49}} & \underline{56.54$\pm${1.57}} \\
& SelfMoA [arXiv'25]  & 46.16$\pm${6.18}    & 49.62$\pm${8.40}    & 54.75$\pm${9.84}    & 90.35$\pm${5.52}    & 31.34$\pm${9.91}    \\
& MoA [ICLR'25]       & 47.33$\pm${6.10}    & 51.32$\pm${3.07}    & 57.85$\pm${2.09}    & 90.67$\pm${2.49}    & 34.08$\pm${4.42}    \\\cline{2-7}
& Ours                & \textbf{69.87$\pm${1.93}}    & \textbf{70.08$\pm${1.81}}    & \textbf{71.10$\pm${1.74}}    & \textbf{92.03$\pm${0.49}}    & \textbf{60.08$\pm${2.55}}   \\
\hline\hline
\end{tabular}
}
\label{tab: comp_nejmqa} %
\end{table*}

\begin{table*}[]
\centering
\caption{Comprehensive evaluations with five evaluation metrics on MMLUPH, demonstrating substantial performance improvements with our method in medical decision-making scenarios.}
\resizebox{1.98\columnwidth}{!}{
\begin{tabular}{l|l|lllll}
\hline\hline
Type                          & LLMs                 & ACC           & F1            & PRE           & SPE           & MCC           \\
\hline
\textcolor{black}{\multirow{4}{*}{Closed-source} }& \textcolor{black}{ Claude Haiku 4.5                   }& \textcolor{black}{ 68.77$\pm$1.85          }& \textcolor{black}{ 68.74$\pm$1.83          }& \textcolor{black}{ 68.99$\pm$1.83          }& \textcolor{black}{ 96.62$\pm$0.01          }& \textcolor{black}{ 65.18$\pm$2.06}          \\
& \textcolor{black}{ GPT-5.1                            }& \textcolor{black}{ 71.35$\pm$3.87          }& \textcolor{black}{ 71.45$\pm$3.87          }& \textcolor{black}{ 72.25$\pm$3.59          }& \textcolor{black}{ 96.79$\pm$0.44          }& \textcolor{black}{ 68.09$\pm$4.32}          \\
& \textcolor{black}{ Gemini 2.5 Flash                   }& \textcolor{black}{ \underline{77.34$\pm$0.67}    }& \textcolor{black}{ \underline{77.34$\pm$0.68}    }& \textcolor{black}{ \underline{77.55$\pm$0.70}    }& \textcolor{black}{ \underline{97.47$\pm$0.08}    }& \textcolor{black}{ \underline{74.77$\pm$0.76}}    \\
& \textcolor{black}{ Ours w/   GPTs }& \textcolor{black}{ \textbf{78.48$\pm$0.44} }& \textcolor{black}{ \textbf{78.47$\pm$0.45} }& \textcolor{black}{ \textbf{78.72$\pm$0.45} }& \textcolor{black}{ \textbf{97.60$\pm$0.05} }& \textcolor{black}{ \textbf{76.05$\pm$0.49}} \\
\hline
\multirow{19}{*}{Open-access} & Medllama2 7B         & 38.96$\pm$7.56    & 41.45$\pm$5.69    & 53.08$\pm$0.93    & 93.40$\pm$1.10    & 34.44$\pm$6.58    \\
& Med42 8B             & 44.84$\pm$1.79    & 45.44$\pm$1.75    & 48.82$\pm$1.27    & 94.03$\pm$0.25    & 38.82$\pm$1.90    \\
& HuatuoGPT-o1 8B      & 32.49$\pm$19.96   & 32.91$\pm$19.89   & 44.03$\pm$10.85   & 92.68$\pm$2.45    & 27.00$\pm$20.35   \\
& {Phi4 14B}             & \underline{70.46$\pm$0.99}    & \underline{70.40$\pm$0.98}    & \underline{71.17$\pm$0.79}    & \underline{96.69$\pm$0.11}    & \underline{67.07$\pm$1.09}    \\
& Qwen2.5 14B          & 62.00$\pm$0.28    & 62.00$\pm$0.24    & 62.40$\pm$0.21    & 95.75$\pm$0.03    & 57.62$\pm$0.33    \\
& Qwen2.5 32B          & 67.77$\pm$0.35    & 67.80$\pm$0.38    & 68.08$\pm$0.41    & 96.40$\pm$0.04    & 64.08$\pm$0.40    \\
& QWQ 32B              & 67.62$\pm$1.10    & 67.76$\pm$0.99    & 68.37$\pm$0.87    & 96.50$\pm$0.09    & 63.94$\pm$1.18    \\
& Openthinker 32B      & 67.81$\pm$0.49    & 67.91$\pm$0.53    & 68.49$\pm$0.57    & 96.52$\pm$0.19    & 64.14$\pm$0.55    \\
& Deepseek-R1 32B      & 62.71$\pm$2.98    & 63.29$\pm$2.26    & 65.01$\pm$0.77    & 95.99$\pm$0.09    & 58.64$\pm$3.07    \\
& Llama3 instruct 70B  & 66.06$\pm$1.84    & 66.02$\pm$1.82    & 66.30$\pm$1.88    & 96.22$\pm$0.21    & 62.18$\pm$2.06    \\
& Qwen1.5 Chat 72B     & 15.29$\pm$0.57    & 13.19$\pm$1.19    & 48.32$\pm$5.27    & 90.64$\pm$0.06    & 12.05$\pm$0.58    \\
& Qwen1.5 Chat 110B    & 45.36$\pm$2.72    & 48.73$\pm$1.49    & 61.13$\pm$2.17    & 93.94$\pm$0.29    & 42.06$\pm$2.10    \\
& dbrx-instruct 132B   & 44.42$\pm$2.37    & 46.14$\pm$2.65    & 51.93$\pm$2.53    & 93.83$\pm$0.28    & 39.11$\pm$2.76    \\
& Mixtral 8x22 141B    & 59.05$\pm$3.23    & 59.10$\pm$3.12    & 60.31$\pm$2.11    & 95.43$\pm$0.36    & 54.39$\pm$3.52    \\
& WizardLM 8x22 141B   & 60.68$\pm$8.89    & 61.19$\pm$7.94    & 64.11$\pm$3.91    & 95.61$\pm$0.98    & 56.57$\pm$9.23    \\
& Debate [ICML'23]     & 67.93$\pm$0.86 & 67.95$\pm$0.83 & 69.49$\pm$0.81 & 96.40$\pm$0.10 & 64.33$\pm$0.95 \\
& SelfMoA   [arXiv'25] & 60.19$\pm$11.26   & 60.86$\pm$10.04   & 63.34$\pm$6.36    & 95.56$\pm$1.25    & 55.99$\pm$11.90   \\
& {MoA [ICLR'25]}        & 65.31$\pm$7.26    & 65.57$\pm$6.76    & 66.98$\pm$5.02    & 96.12$\pm$0.81    & 61.38$\pm$7.98    \\ \cline{2-7}
 & {Ours}                 & \textbf{71.35$\pm$3.87}    & \textbf{71.45$\pm$3.87}    & \textbf{72.25$\pm$3.59}    & \textbf{96.79$\pm$0.44}    & \textbf{68.09$\pm$4.32}   \\
\hline\hline
\end{tabular}
}
\label{tab: comp_mmluph} %
\end{table*}

\begin{figure}
    \centering
    \includegraphics[width=0.48\textwidth]{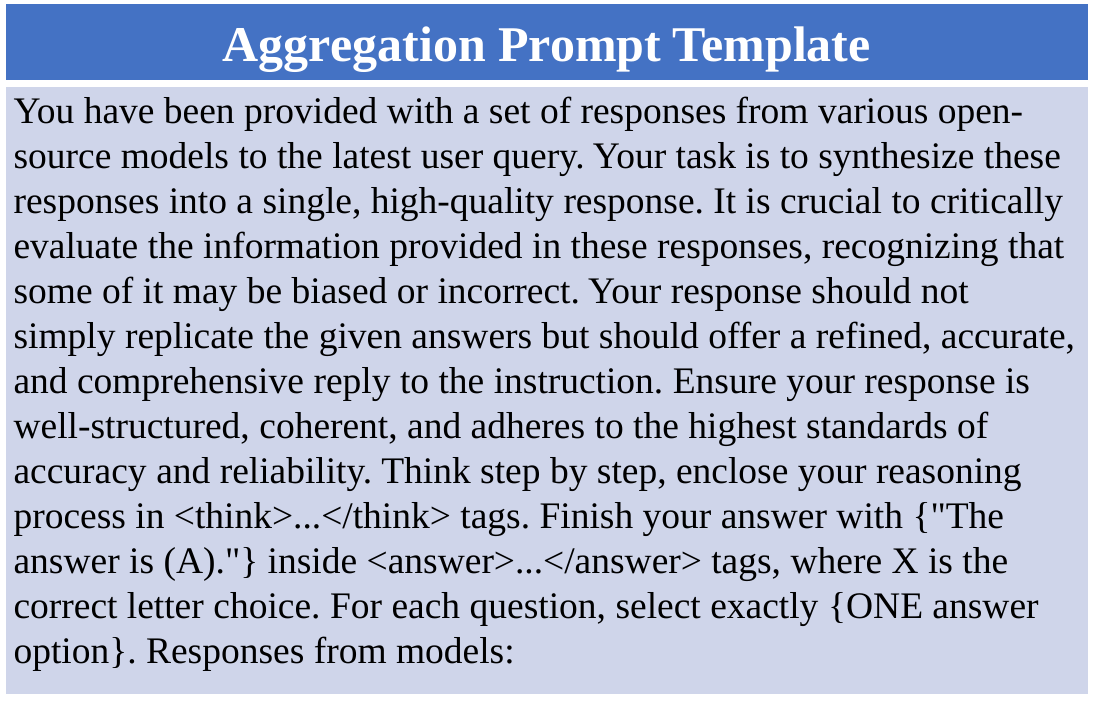}
    \vspace{0.1em}
    \includegraphics[width=0.48\textwidth]{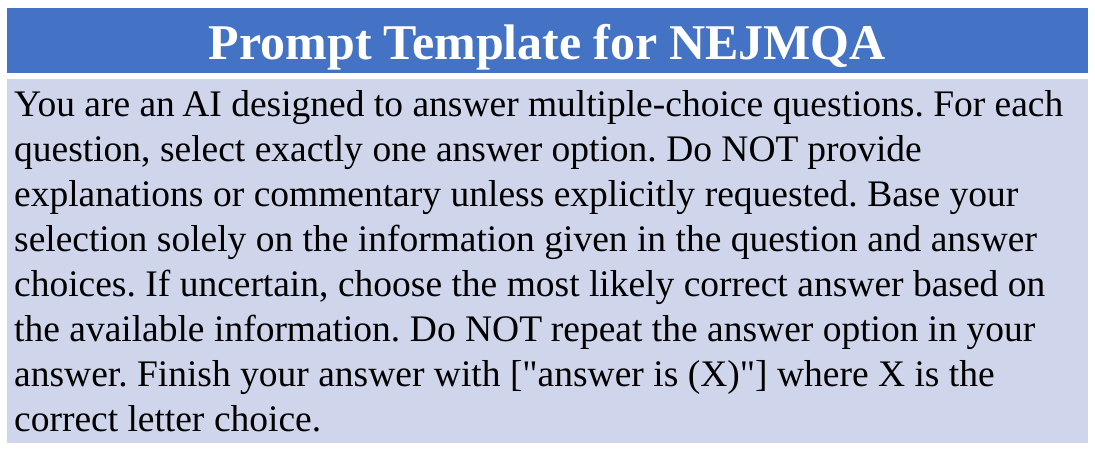}
    \vspace{0.6em}
    \includegraphics[width=0.48\textwidth]{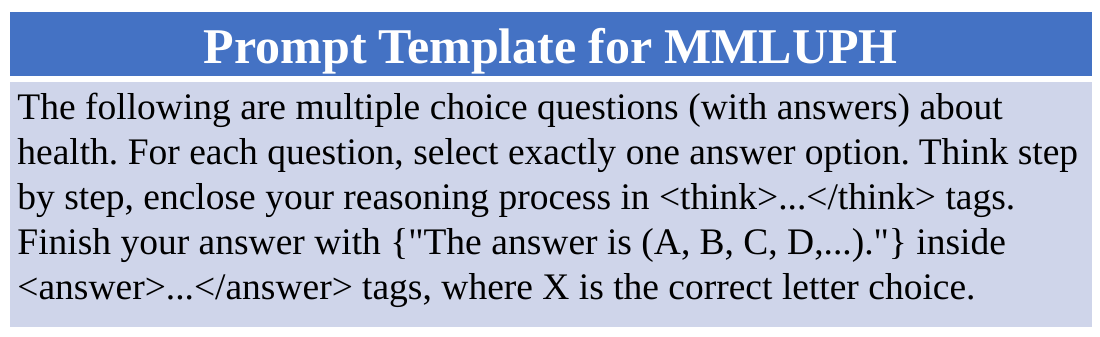}
    \vspace{0.5em}
    \includegraphics[width=0.48\textwidth]{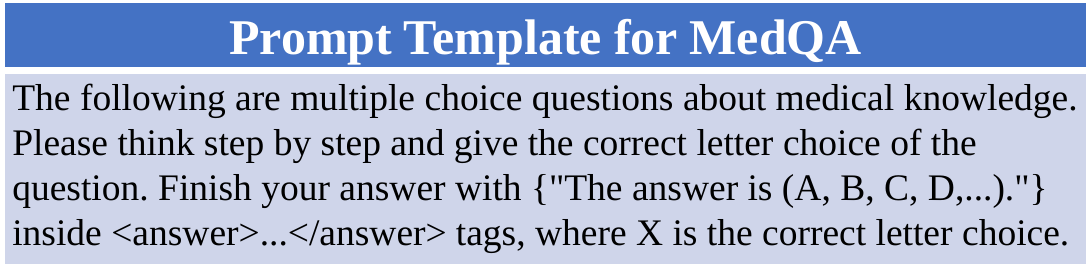}
    \caption{Specified prompts for output aggregation and used datasets.}
    \label{fig: prompt}
\end{figure}

For fair comparisons, we follow the same prompt template setting as \cite{du2023improving,wang2025mixtureofagents,li2024rethinking} to conduct the aggregation of output and the temperature parameter of all LLMs is set to the default value of $0.7$. Fig.~\ref{fig: prompt} shows the template of the aggregate-and-synthesize prompt to integrate responses of other models, where the original user query is included immediately after this system prompt. In the experiments, we heuristically mask two agents in each layer until only one agent is used to achieve the final inference, more details could be found in Sec. \ref{eprm: analysis}. We test these open-access LLMs through the Ollama platform \cite{ollama2023} and the closed-source LLMs via APIs through OpenAI. The results were reported as the mean values and their corresponding standard deviations, obtained by repeating the experiments three times, represented as mean $\pm$ std in the experimental results. The model is implemented with PyTorch on NVIDIA GeForce RTX 4090 and NVIDIA RTX A6000. We ensure strict adherence to the licensing terms of all LLMs utilized in this work. 

\noindent \textbf{Metrics.} To comprehensively evaluate the performance of the compared models and our method, we exploit a series of evaluation metrics, including accuracy (ACC), weighted F1-score (F1), Precision (PRE) \cite{powers2020evaluation}, Specificity (SPE) \cite{saah1998sensitivity}, and Matthews Correlation Coefficient (MCC) \cite{matthews1975comparison}.

\subsection{{Compared Results}}

\begin{table*}[]
\centering
\caption{Comprehensive evaluations with five evaluation metrics on MedQA, demonstrating substantial performance improvements with our method in medical decision-making scenarios.}
\resizebox{1.98\columnwidth}{!}{
\begin{tabular}{l|l|lllll}
\hline\hline
Type                          & LLMs                & ACC                         & F1                          & PRE                         & SPE                         & MCC                         \\
\hline
\textcolor{black}{\multirow{4}{*}{Closed-source}} & \textcolor{black}{Claude Haiku 4.5}                   & \textcolor{black}{71.93$\pm$0.36}          & \textcolor{black}{72.20$\pm$0.35}          & \textcolor{black}{73.22$\pm$0.78}          & \textcolor{black}{94.21$\pm$0.09}          & \textcolor{black}{65.14$\pm$0.60}          \\
& \textcolor{black}{GPT-5.1                            }& \textcolor{black}{\underline{91.04$\pm$0.57}    }& \textcolor{black}{\underline{91.08$\pm$0.54}    }& \textcolor{black}{\underline{91.18$\pm$0.49}    }& \textcolor{black}{\underline{98.11$\pm$0.25}    }& \textcolor{black}{\underline{88.79$\pm$0.70}}    \\
& \textcolor{black}{Gemini 2.5 Flash                   }& \textcolor{black}{90.18$\pm$0.49          }& \textcolor{black}{90.18$\pm$0.49          }& \textcolor{black}{90.21$\pm$0.49          }& \textcolor{black}{97.53$\pm$0.12          }& \textcolor{black}{87.69$\pm$0.61}          \\
& \textcolor{black}{\textcolor{black}{Ours w/ GPTs} }& \textcolor{black}{\textbf{92.59$\pm$0.60} }& \textcolor{black}{\textbf{92.59$\pm$0.60} }& \textcolor{black}{\textbf{92.62$\pm$0.59} }& \textcolor{black}{\textbf{98.14$\pm$0.15} }& \textcolor{black}{\textbf{90.71$\pm$0.75}} \\
\hline
\multirow{19}{*}{Open-access} & Medllama2 7B        & 29.98$\pm${16.91}         & 31.36$\pm${14.78}         & 36.31$\pm${16.85}         & 88.81$\pm${1.32}          & 16.65$\pm${18.48}         \\
& Med42 8B            & 55.96$\pm${2.77}          & 55.91$\pm${2.75}          & 56.13$\pm${2.75}          & 90.22$\pm${0.41}          & 44.78$\pm${3.47}          \\
& HuatuoGPT-o1 8B     & 27.09$\pm${30.63}         & 31.44$\pm${27.09}         & 55.11$\pm${8.46}          & 89.98$\pm${0.54}          & 24.19$\pm${25.31}         \\
& Phi4 14B            & 75.15$\pm${1.01}          & 75.24$\pm${1.03}          & 75.42$\pm${1.08}          & 95.08$\pm${0.33}          & 68.88$\pm${1.28}          \\
& Qwen2.5 14B         & 64.55$\pm${1.22}          & 64.60$\pm${1.14}          & 64.93$\pm${1.15}          & 92.14$\pm${0.79}          & 55.67$\pm${1.51}          \\
& Qwen2.5 32B         & 66.11$\pm${2.16}          & 66.27$\pm${2.10}          & 66.99$\pm${1.71}          & 93.90$\pm${0.59}          & 57.71$\pm${2.63}          \\
& QWQ 32B             & 78.76$\pm${0.37}          & 79.05$\pm${0.33}          & 79.43$\pm${0.25}          & \underline{96.17$\pm${0.59}}          & 73.45$\pm${0.45}          \\
& Openthinker 32B     & 77.63$\pm${1.35}          & 77.96$\pm${1.18}          & 78.42$\pm${1.06}          & 96.00$\pm${0.38}          & 72.05$\pm${1.65}          \\
& Deepseek-R1 32B     & 76.98$\pm${0.36}          & 77.70$\pm${0.27}          & 78.59$\pm${0.57}          & 96.07$\pm${0.67}          & 71.35$\pm${0.42}          \\
& Llama3 instruct 70B & 71.41$\pm${1.16}          & 71.37$\pm${1.31}          & 71.49$\pm${1.52}          & 93.22$\pm${0.96}          & 64.18$\pm${1.50}          \\
& Qwen1.5 Chat 72B    & 29.51$\pm${18.39}         & 34.94$\pm${13.87}         & 50.93$\pm${0.86}          & 90.22$\pm${0.39}          & 23.45$\pm${13.00}         \\
& Qwen1.5 Chat 110B   & 58.42$\pm${3.30}          & 60.87$\pm${1.57}          & 63.88$\pm${1.62}          & 93.57$\pm${1.29}          & 49.24$\pm${3.10}          \\
& dbrx-instruct 132B  & 51.20$\pm${6.47}          & 52.67$\pm${4.73}          & 56.04$\pm${1.37}          & 90.08$\pm${1.03}          & 40.24$\pm${6.45}          \\
& Mixtral 8x22 141B   & 60.88$\pm${1.31}          & 63.12$\pm${0.89}          & 65.85$\pm${1.87}          & 93.67$\pm${1.37}          & 52.12$\pm${1.17}          \\
& WizardLM 8x22 141B  & 70.04$\pm${2.41}          & 71.67$\pm${1.00}          & 73.62$\pm${0.61}          & 94.55$\pm${0.18}          & 63.08$\pm${2.49}          \\
&  Debate [ICML'23]    & \underline{79.44$\pm${0.25}}       & \underline{79.47$\pm${0.26}}       & \underline{80.38$\pm${0.48}}       & 94.83$\pm${0.06}       & \underline{74.39$\pm${0.34}}       \\
& SelfMoA [arXiv'25]  & 47.96$\pm${11.93}         & 55.64$\pm${5.91}          & 71.66$\pm${7.10}          & 92.70$\pm${0.51}          & 42.19$\pm${8.91}          \\
& MoA [ICLR'25]       & 57.06$\pm${2.07}          & 61.27$\pm${2.53}          & 67.17$\pm${6.30}          & 93.22$\pm${1.56}          & 48.65$\pm${1.88}          \\\cline{2-7}
& Ours                & \textbf{79.94}$\pm${\textbf{1.31}} & \textbf{80.24}$\pm${\textbf{1.21}} & \textbf{80.66}$\pm${\textbf{1.03}} & \textbf{97.01}$\pm${\textbf{0.21}} & \textbf{74.94}$\pm${\textbf{1.60}} \\
\hline\hline
\end{tabular}
}
\label{tab: comp_medqa} %
\end{table*}

\noindent \textbf{Multiple Metrics Evaluation.} 
We conduct the experimental results with five evaluation metrics on three medical decision-making benchmarks. As shown in Tables \ref{tab: comp_nejmqa}, \ref{tab: comp_mmluph}, and \ref{tab: comp_medqa}, we have the following observations:
\begin{itemize} 
    \item Our framework, composed of 14B to 32B open-access LLMs, can exceed the framework composed of 70B to 141B open-access LLMs, indicating that the adaptive progressive propagation is capable of improving the medical decision-making capability of the MAS-based model. It benefits from the innovative adaptive progressive propagation and the cross-consistency maximization mechanism that ensures enhanced consistency and superior medical decision-making.
    \item Our framework outperforms closed-source LLM Claude Haiku 4.5 among all comparisons, demonstrating its effectiveness even with modestly sized open-access foundation models. For example, on MedQA, our approach improves 8.01\% over Claude Haiku 4.5 on ACC, 8.04\% on F1, 7.44\% on PRE, 2.80\% on SPE, and 9.80\% on MCC. 
    \item These advancements highlight the accelerating momentum in open-access research, where innovative architectural designs are rapidly closing the performance gap with proprietary LLM (e.g., closed-source model GPT-5.1), paving the way for more accessible and powerful medical decision-making tools in the near future.
\end{itemize}

\begin{figure}[t]
    \centering

          \includegraphics [width=0.98 \linewidth]{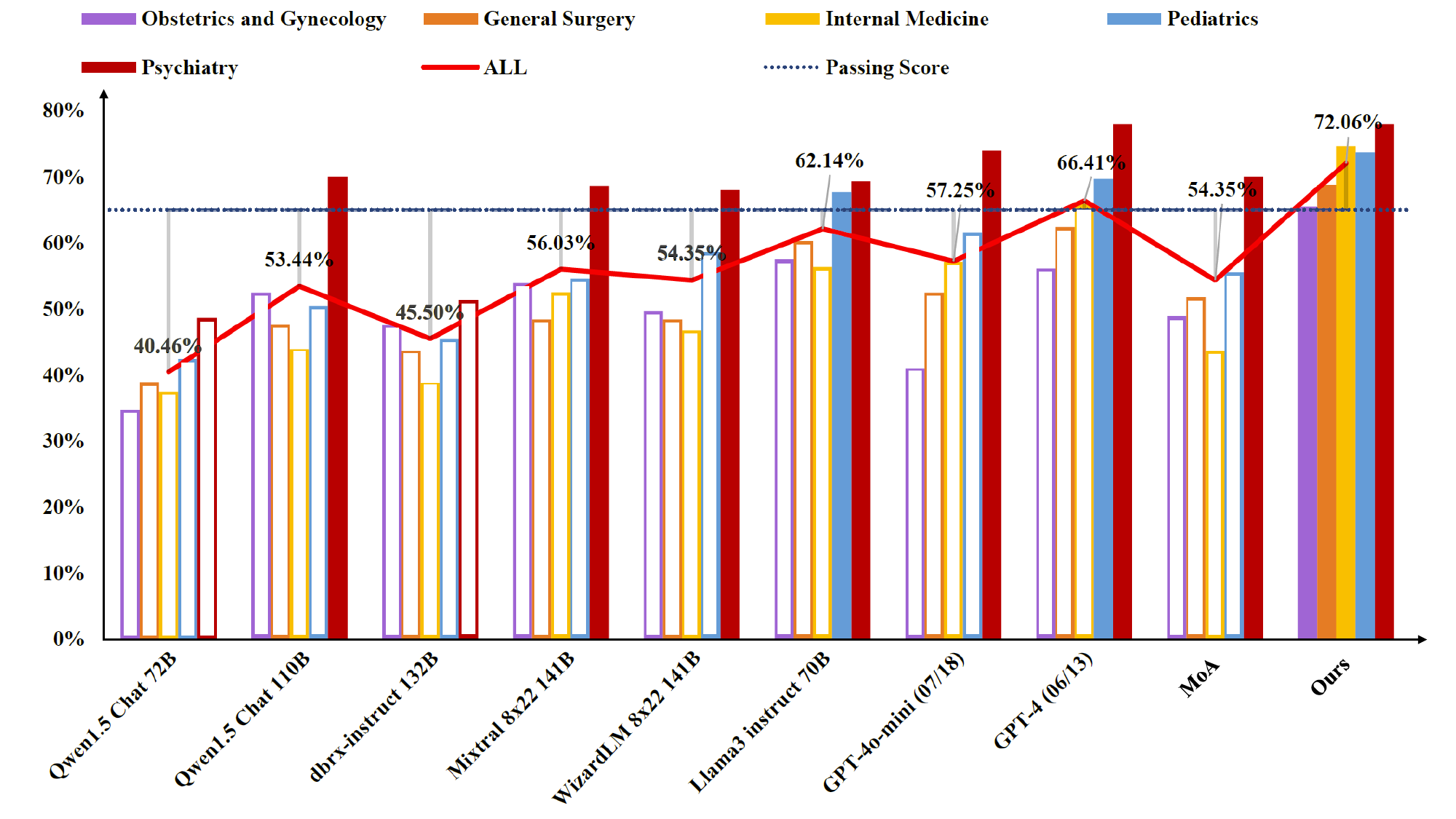}
    \caption{Evaluation across diverse disciplines on NEJMQA.} 
    \label{fig: diverse disc}
\end{figure}

\begin{figure}[!]
    \centering
    \includegraphics [width=0.98 \linewidth]{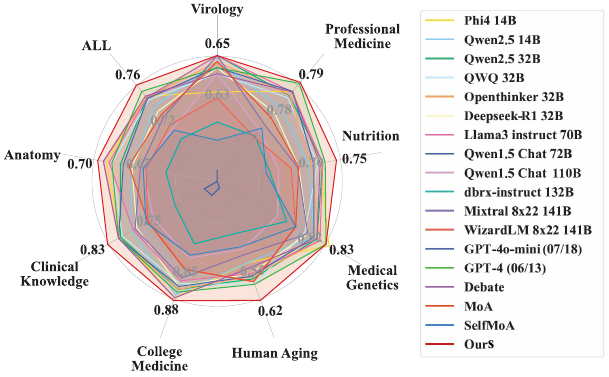}
    \caption{Comparisons across diverse disciplines on MMLUPH.}
    \label{fig: all_comp}
\end{figure}

\noindent\textbf{Diverse Disciplines Comparisons.}
Moreover, to assess the medical decision-making capacity of LLMs, we conduct the comparisons on NEJMQA in five clinical disciplines (Fig. \ref{fig: diverse disc}) and  MMLUPH in eight medical specialties (Fig.~\ref{fig: all_comp}). As shown in Fig. \ref{fig: diverse disc}, most individual LLMs are uncompetitive on a medical specialist licensing examination, falling below the minimum passing score of 65\%, whereas our MAC framework boosts LLM medical decision-making capacity and demonstrates substantial performance improvements across diverse disciplines, with the model that reaches the passing score highlighted by filling its shape. Furthermore, the performance of MoA \cite{wang2025mixtureofagents} is worse than that of some individual LLMs in terms of overall ACC, indicating that the multi-agent collaboration, which performs well in general AI tasks, does not work in medical decision-making scenarios. Although GPT-4 achieves 66.41\% in overall performance, it does not reach the official passing score of 65\% in \textit{Obstetrics and Gynecology} and \textit{General Surgery} disciplines, indicating that even advanced closed-source models still have a gap in medical decision-making scenarios. In contrast, our method obtains the best performance on NEJMQA and MMLUPH across physician-oriented disciplines, verifies the effectiveness of the proposed framework.

\noindent\textbf{Inference Cost Analysis.}
We analyze the inference costs of running time and memory occupation relative to ACC performance in Fig. \ref{fig: motivation}, where we observe that most individual LLMs show insignificant performance improvements due to limitations in their model size. Although MoA surpasses most individual LLMs through iterative aggregation of model outputs, it incurs substantial memory occupation and running time. In contrast, our framework achieves over 6.11\% higher ACC than MoA while reducing memory usage by 70,206 MB and running time by 41,855.52 seconds.

\subsection{Experimental Analyses}
\label{eprm: analysis}
We conduct ablation studies to evaluate the effectiveness of Pareto-optimal agent construction (PAC) and cross-consistency maximization (CCM) strategies and further analyze the influence of different masked mechanisms.

\begin{table}[]
\centering
\caption{The ablation study of the proposed PAC and CCM term on NEJMQA, where {\XSolidBrush} and {\Checkmark} in each row indicates the non-use or use of the corresponding component, respectively.}
\resizebox{0.96\columnwidth}{!}{
\begin{tabular}{cc|ccccc}
\hline\hline
PAC                            & CCM                            & ACC             & F1              & PRE             & SPE             & MCC                           \\ \hline\hline
{\XSolidBrush} & {\XSolidBrush} & 54.35           & 54.82           & 55.62           & 87.85           & 39.15                      \\
                        {\XSolidBrush} & {\Checkmark} & 62.60           & 62.75           & 63.24           & 90.04           & 50.08                      \\
                        {\Checkmark} & {\XSolidBrush} & \underline{65.95} & \underline{66.83} & \underline{68.25} & \underline{91.12} & \underline{55.11}  \\
                        {\Checkmark} & {\Checkmark} & \textbf{72.06}  & \textbf{72.13}  & \textbf{73.11}  & \textbf{92.59}  & \textbf{62.98}   \\ 
                        \hline\hline
\end{tabular}
}
\label{tab: as_sd_cc} %
\end{table}

\textbf{Ablation Study of PAC and CCM Strategies.}
We conduct comprehensive ablation studies with five evaluation metrics to investigate the proposed PAC and CCM strategies. The experimental results are listed in Table \ref{tab: as_sd_cc}, where the first row denotes the baseline \cite{wang2025mixtureofagents}, the second and third row denote the variant of the baseline that exploits the PAC and CCM strategies, respectively, and the fourth row is our method, denoted as Ours. From Table \ref{tab: as_sd_cc}, we have the following observations that the advantage of the PAC and CCM strategies could be validated by comparing the results of the second and third rows with the fourth row of each metric. For example, on NEJMQA, it can be seen that simultaneously considering the PAC and CCM strategies could produce an ACC improvement of 6.11\% to 9.46\%.

\begin{table}[!]
\centering
\caption{{The analyses of the masked agent strategies.}}
\resizebox{0.99\columnwidth}{!}{
\begin{tabular}{c|ccccc}
\hline\hline
Masked Strategy & ACC            & F1             & PRE            & SPE            & MCC            \\ \hline\hline
Baseline & 54.35          & 54.82          & 55.62          & 87.85          & 39.15          \\
                        Random   & 61.22          & 61.33          & 61.53          & 89.66          & 48.17          \\
                        \textcolor{black}{Soft-weighting} & \textcolor{black}{\underline{66.11}}          & \textcolor{black}{64.29}          & \textcolor{black}{62.76}          & \textcolor{black}{\textbf{97.56} }         & \textcolor{black}{\underline{55.93} }         \\
                        Sequence & \underline{66.11}          & \underline{66.22}          & \underline{66.98}          & {91.00}          & {54.93}          \\
                        Ours     & \textbf{72.06} & \textbf{72.13} & \textbf{73.11} & \underline{92.59} & \textbf{62.98} \\ 
                        \hline\hline
\end{tabular}
}
\label{tab: as_mask}
\end{table}


\begin{figure}[t]
    \centering
    \includegraphics[width=0.9\linewidth]{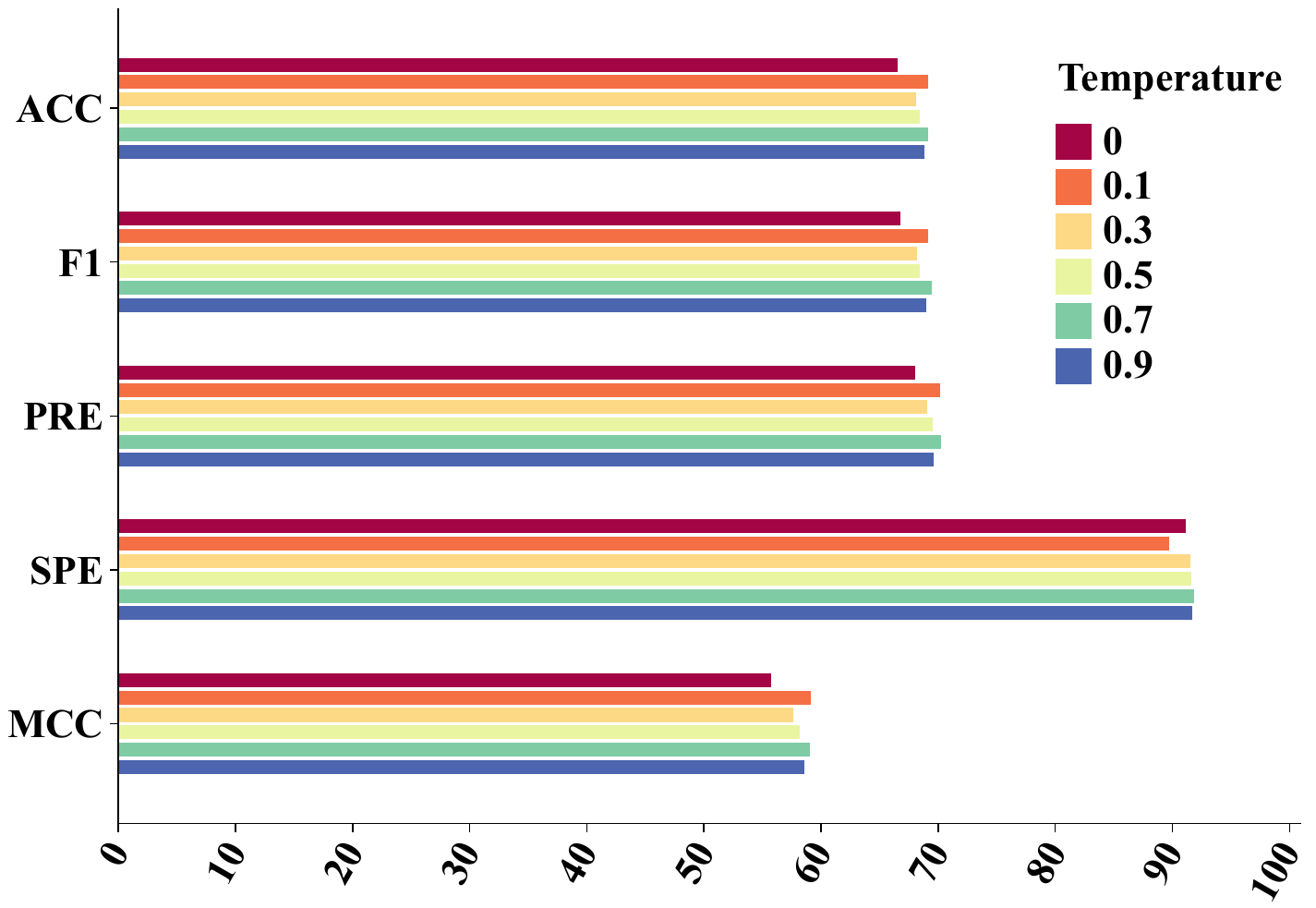}
    \caption{The analyses of different LLM temperature parameters.}
   \label{fig: as_diff_temp}
\end{figure}

\begin{figure}[t]
    \centering
    \includegraphics[width=0.9\linewidth]{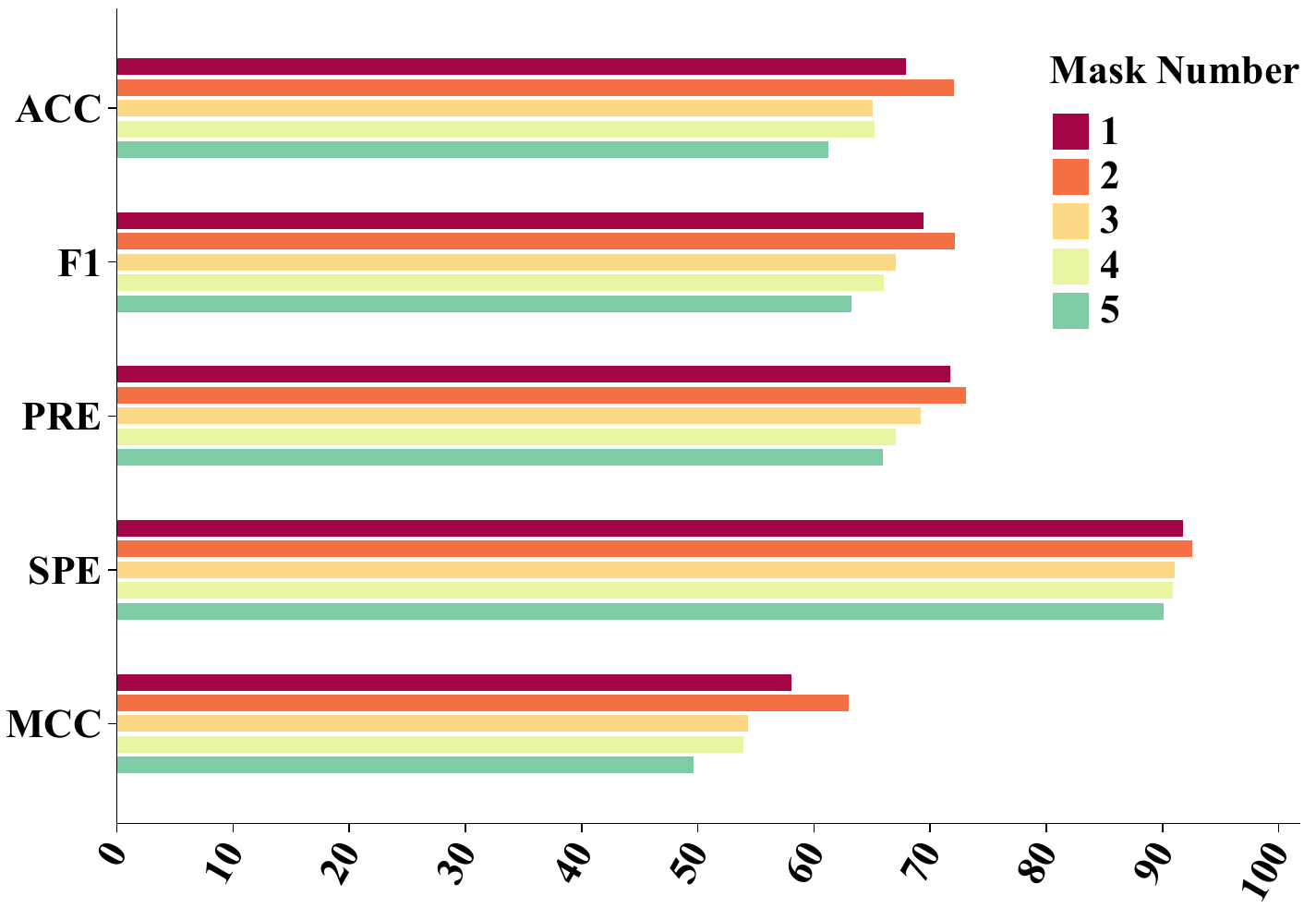}
    \caption{The analyses of the masked number of agents.}
   \label{fig: as_mask_num}
\end{figure}

\begin{figure}[t]
    \centering
        \includegraphics [width=0.68 \linewidth]{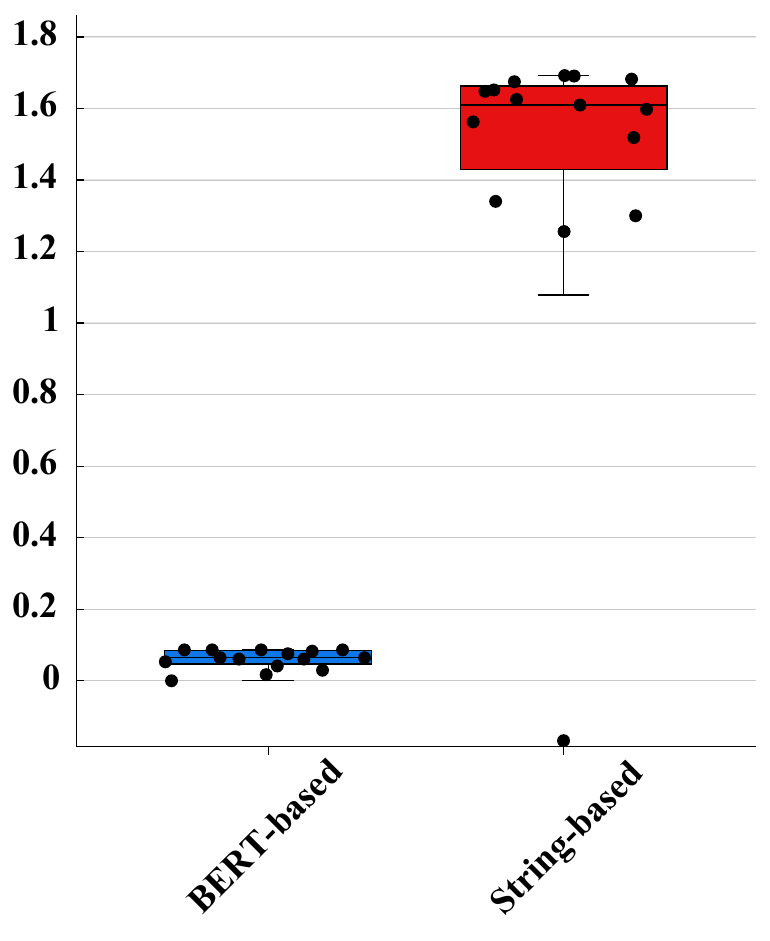}
    \caption{\textcolor{black}{Boxplot of the distribution of BERT-based (left) and string-based (right) metrics towards the diversity scores among different LLMs. The logarithmic function is used to scale the over-distorted boxplot of string-based one.}} 
    \label{fig: boxplot-bert-string}
\end{figure}

\textbf{Analyses of Diversity Calculation.} 
\textcolor{black}{In the preliminary exploration, we did investigate the semantic embedding similarity metric (e.g., BERTScore \cite{zhang2019bertscore}) to calculate the diversity score of LLMs. However, we empirically found that the BERT-based metric failed to provide sufficient discriminability across different LLMs especially in the long medical context scenario, falling into decision paralysis. This may be attributed to the fact that long medical contexts cause a semantic-averaging effect in LLMs, homogenizing sentence embeddings and making it difficult to distinguish structural or subtle output variations. 
We provide a statistical comparison of different metrics (BERT-based vs. string-based) for diversity calculation across various LLMs, as shown in Table \ref{tab: bert-string}. Specifically, the BERT-based metric exhibits a very narrow distribution of diversity scores across fifteen LLMs, only ranging from 1.00 to 1.22. In contrast, the string-based metric shows a much more distinct and wider distribution, ranging from 0.6802 to 49.0705, better differentiating the diversity among different LLMs. To have an intuitive observation, we present a boxplot of the distribution of BERT-based and string-based metrics in Fig. \ref{fig: boxplot-bert-string}, where a logarithmic (LOG10) transformation is applied to mitigate their over-distorted distributions. We can find that string-based metric provides a much higher discriminability since it is capable of strictly capturing the structural and lexical variations in these long medical contexts. 
We strictly force LLMs to output their rationales within the \texttt{<think>...</think>} format and their final decisions in the highly standardized format (\texttt{<answer>} The answer is \{X\}. \texttt{</answer>}, where X is the correct letter choice). Therefore, the string-based matching is capable of inherently bypassing the phrasing variance and ensuring stable consistency calculation.}

\begin{table}[h!]
\centering
\caption{\textcolor{black}{Comparison of different metrics for diversity calculation across various LLMs.}}
\resizebox{0.8\columnwidth}{!}{
\begin{tabular}{c|cc}
\hline\hline
LLMs                & BERT & String \\
\hline
Medllama2 7B        & 1.13 & 49.21  \\
Med42 8B            & 1.15 & 18.03  \\
HuatuoGPT-o1 8B     & 1.04 & 49.07  \\
Phi4 14B            & 1.16 & 40.71  \\
Qwen2.5 14B         & 1.22 & 44.46  \\
Qwen2.5 32B         & 1.16 & 33.02  \\
QWQ 32B             & 1.10 & 44.86  \\
Openthinker 32B     & 1.15 & 48.10  \\
Deepseek-R1 32B     & 1.07 & 47.30  \\
Llama3 instruct 70B & 1.00 & 0.68   \\
Qwen1.5 Chat 72B    & 1.22 & 39.60  \\
Qwen1.5 Chat 110B   & 1.22 & 36.53  \\
dbrx-instruct 132B  & 1.22 & 42.21  \\
Mixtral 8x22 141B   & 1.21 & 19.96  \\
WizardLM 8x22 141B  & 1.19 & 21.90 \\
\hline\hline
\end{tabular}
}
\label{tab: bert-string}
\end{table}

\textbf{Failure Mode Analysis.} 
\textcolor{black}{We have conducted a failure mode analysis with the frequency statistics on NEJMQA to verify our assumption that the majority is correct in most medical decision-making scenarios rather than all cases. As shown in Table \ref{tab: Frequency statistics}, we can find that although the ``lone genius'' scenario (the majority is wrong, the minority is true) does occur, its frequency is extremely low, accounting for only 5 out of 655 (0.76\%) cases. Conversely, the case that the majority consensus is reliable holds true in the vast majority of scenarios (80.15\%), and both wrong (19.09\%). This phenomenon strongly supports our core design rationale that prioritizing the majority consensus yields a highly favorable and robust trade-off for overall accuracy.}

\begin{table}[t]
\centering
\caption{\textcolor{black}{Frequency statistics of failure modes on NEJMQA. "Major" and "Minor" indicate the majority consensus and minority views, respectively.}}
\resizebox{0.6\columnwidth}{!}{
\begin{tabular}{cc|c}
\hline\hline
Major & Minor & Frequency \\
\hline
{\XSolidBrush}     & {\XSolidBrush}          & 125       \\
{\XSolidBrush}     & {\Checkmark}         & 5         \\
{\Checkmark}     & {\XSolidBrush}          & 187       \\
{\Checkmark}     & {\Checkmark}         & 338         \\
\hline\hline
\end{tabular}
}
\label{tab: Frequency statistics}
\end{table}

\textbf{Analyses of Temperature Parameter Values.} 
The temperature parameter is used to control the output randomness of an LLM by modifying the probability distribution of the next token. We evaluate the influence of our MAC model with respect to different temperature parameter values on five metrics: ACC, F1, PRE, SPE, and MCC, as shown in Fig. \ref{fig: as_diff_temp}. We have the following observations that our model is relatively robust to the temperature parameter, which is capable of achieving almost the same performance in terms of the five metrics in a wide range of $\mathcal{S}=\left\{ 0, 0.1, 0.3, 0.5, 0.7, 0.9 \right\}$, illustrating that LLM is robust to the temperature parameter. In this work, we set it as $0.7$.

\textbf{Analysis of The Masked Agent Strategy and Masked Number.}
To evaluate the advantages of the CC-driven MAC framework, we analyze the impact of different masked agent strategies in Table \ref{tab: as_mask}. The ‘baseline’ refers to layers without a masked agent strategy, where all agents participate in the aggregation. The ‘random’ strategy randomly masks agents layer by layer. \textcolor{black}{The ‘soft-weighting’ strategy \cite{chen2024reconcile} has each agent output a prediction and self-rated confidence, then aggregates predictions weighted by confidence.} The ‘sequence’ strategy masks the agent with the lowest individual diversity score layer by layer in ascending order. The ‘Ours’ strategy denotes the proposed CC-driven masked agent strategy. Table \ref{tab: as_mask} shows that the usage of the CCM mechanism to adaptively mask low-consistency agents in each layer is effective in improving the medical decision-making capability. \textcolor{black}{Moreover, retaining all responses via soft-weighting introduces a 25\% computational overhead (324,190s vs 260,668s) and severely degrades ACC from 72.06\% to 66.11\%.}

Additionally, we evaluate the effect of varying the number of masked agents, as given in Fig. \ref{fig: as_mask_num}. For those settings where the number of agents in the last layer is not 1, we use Openthinker 32B, which is the same as the LLM used in the last layer of our framework, to perform an aggregation operation for fair comparisons.  The results indicate that setting the number of masked agents to 2 could significantly eliminate inconsistent outputs to ensure robust collaboration.

\textbf{\textcolor{black}{Analyses of Open-ended Medical Scenarios.}} 
\textcolor{black}{
To demonstrate our robust reasoning capabilities beyond multiple-choice constraints, we extend the evaluation to two open-ended medical scenarios: Medications \cite{abacha2019bridging}, which consists of 690 drug-related consultations with an average question and answer length of 10.03 and 90.96 tokens respectively; and MedQA-open \cite{nachane2024few}, an open-ended reformulation of MedQA containing 1,155 samples with significantly longer contexts (averaging 164.05 tokens per question). The results are demonstrated in Tables \ref{tab:Medications} and \ref{tab:MedQA-open}. We prioritize evaluating the clinical quality of model responses using GPT-5.4 across medically meaningful criteria \cite{abacha2019bridging,nachane2024few,tu2025towards}: Clinical Correctness (CCor), Key Point Coverage (KPC), Management Appropriateness (MA), Clarity and Actionability (CA), and clinically Acceptable Rate (AR). 
As shown in Table \ref{tab:Medications}, our framework effectively handles practical drug-related queries, achieving the highest clinically Acceptable Rate (AR) of 70.33\% and obtaining the best or tied-best performance on critical dimensions including CCor (1.24), KPC (1.36), MA (1.32), and CA (1.91). Similarly, in Table \ref{tab:MedQA-open}, our framework again delivers the best overall clinical performance, achieving the highest scores on CCor (1.32), KPC (1.44), MA (1.30), CA (1.90), and AR (68.67\%). These substantial improvements on clinical metrics explicitly demonstrate that our CC masking and adaptive propagation mechanisms genuinely aggregate complementary clinical knowledge to resolve complex open-ended scenarios, rather than merely exploiting substring-level agreement.}


\begin{table}[t]
\centering
\caption{\textcolor{black}{Performance evaluation on Medications dataset.}}
\resizebox{0.96\columnwidth}{!}{
\begin{tabular}{c|ccccc}
\hline\hline
Model           & CCor            & KPC           & MA            & CA            & AR             \\
\hline
Phi4 14B        & 1.20          & 1.13          & 1.21          & 1.69          & 67.00          \\
Qwen2.5 14B     & 1.19          & 1.05          & 1.19          & 1.74          & \underline{67.67}    \\
Qwen2.5 32B     & 0.00          & 0.00          & 0.00          & 0.25          & 0.00           \\
QWQ 32B         & 0.00          & 0.00          & 0.01          & 0.26          & 0.00           \\
Openthinker 32B & \textbf{1.24} & \underline{1.25}    & \underline{1.24}    & \underline{1.83}    & 67.33          \\
Deepseek-R1 32B & \underline{1.21}    & 1.12          & 1.17          & 1.77          & 67.33          \\
Ours            & \textbf{1.24} & \textbf{1.36} & \textbf{1.32} & \textbf{1.91} & \textbf{70.33} \\
\hline\hline
\end{tabular}
}
\label{tab:Medications}
\end{table}

\begin{table}[t]
\centering
\caption{\textcolor{black}{Performance evaluation on MedQA-open dataset.}}
\resizebox{0.96\columnwidth}{!}{
\begin{tabular}{c|ccccc}
    \hline\hline
    Model           & CCor            & KPC           & MA            & CA            & AR             \\ 
    \hline
    Phi4 14B        & \underline{1.31}    & 1.28          & 1.21          & 1.84          & 64.67          \\
    Qwen2.5 14B     & 0.99          & 1.03          & 1.09          & 1.71          & 48.33          \\
    Qwen2.5 32B     & 1.10          & 1.08          & 1.12          & 1.72          & 56.00          \\
    QWQ 32B         & 1.25          & 1.41          & 1.23          & 1.87          & \underline{67.00}    \\
    Openthinker 32B & 1.26          & \underline{1.42}    & \underline{1.24}    & \underline{1.89}    & 66.33          \\
    Deepseek-R1 32B & 1.17          & 1.25          & 1.17          & 1.88          & 60.00          \\
    Ours            & \textbf{1.32} & \textbf{1.44} & \textbf{1.30} & \textbf{1.90} & \textbf{68.67} \\
    \hline\hline
\end{tabular}
}
\label{tab:MedQA-open}
\end{table}

\textbf{\textcolor{black}{Analyses of Generalizability.}} 
\textcolor{black}{To reflect both profound clinical motivation and broader generalizability of the proposed MAC, we conduct the evaluation on a general-domain mathematical reasoning task, as shown in Table \ref{tab:MMLU-Pro-Math}. Specifically, the evaluation is conducted on the MMLU-Pro Math dataset, which comprises 1,351 questions with 10 answer options for each question. Crucially, complex medical diagnostic pathways (e.g., step-by-step differential diagnosis) and rigorous deduction chains in mathematical proofs share a profound algorithmic isomorphism. Both paradigms inherently rely on sequential logical aggregation and the iterative pruning of contradictory reasoning branches. The substantial improvements confirm that our framework not only addresses the stringent requirements of medical decision-making but also exhibits strong applicability to general complex reasoning scenarios. }

\begin{table}[!]
    \centering
    \caption{\textcolor{black}{Performance comparisons on MMLU-Pro Math.}}
    \resizebox{0.98\columnwidth}{!}{
    \begin{tabular}{c|ccccc}
    \hline\hline
        LLMs            & ACC            & F1             & PRE            & SPE            & MCC            \\
        \hline
        Phi4 14B        & 79.42          & 80.90          & 82.79          & 97.95          & 77.28          \\
        Qwen2.5 14B     & 73.95          & 75.07          & 76.60          & 97.39          & 71.17          \\
        Qwen2.5 32B     & 76.68          & 77.88          & 79.37          & 97.67          & 74.20          \\
        QWQ 32B         & 85.12          & \underline{89.24}    & \textbf{93.81} & 98.59          & 84.06          \\
        Openthinker 32B & 84.83          & 88.34          & 92.31          & 98.55          & 83.60          \\
        Deepseek-R1 32B & 84.90          & 88.45          & 92.40          & 98.56          & 83.69          \\
        Debate          & \underline{86.75}    & 88.19          & 89.92          & \underline{98.69}    & \underline{85.38}    \\
        Ours            & \textbf{89.79} & \textbf{91.52} & \underline{93.47}    & \textbf{99.00} & \textbf{88.77} \\
        \hline\hline
    \end{tabular}
    }
    \label{tab:MMLU-Pro-Math}
\end{table}



\begin{figure*}[!]
    \centering
    \includegraphics [width=0.92 \linewidth]{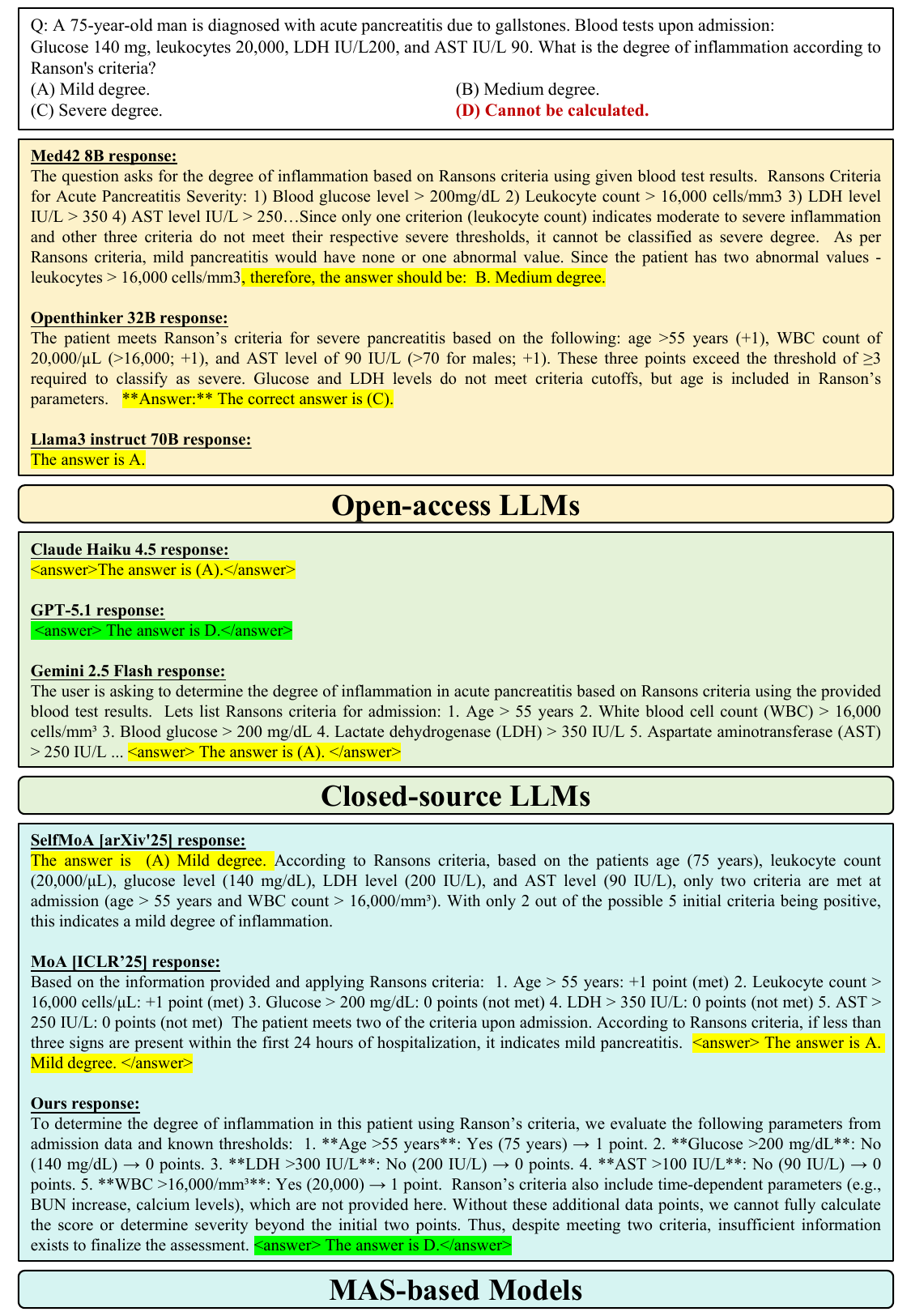}
    \caption{Case study analysis of open-access, closed-source LLMs, and MAS-based models on clinical question-answering.} 
    \label{fig: casestudy}
\end{figure*}

\noindent\textbf{Case Study.} 
In this section, we conduct a comprehensive case study analysis of open-access, closed-source LLMs, and MAS-based models on clinical question-answering, as shown in Fig. \ref{fig: casestudy}. The evaluations reveal that state-of-the-art models demonstrate frequent inaccuracies, such as incomplete application of the criteria (focusing solely on age and WBC while ignoring others), and premature classification of severity without acknowledging the need for 48-hour parameters to compute the full score. In contrast, our model excels by precisely recalling all five admission criteria thresholds, accurately assigning points, and critically recognizing that Ranson's full prognostic score requires additional 48-hour data (e.g., hematocrit drop, BUN rise, calcium levels, base deficit, PaO2, and fluid sequestration), leading to the correct conclusion that the degree cannot be fully calculated with the given information. These findings underscore that, although leading models possess substantial capabilities, their application in high-stakes medical domains necessitates rigorous validation and iterative enhancements to address distortions in factual medical knowledge and lapses in data completeness verification, thereby promoting consistency, interpretability, and reliability in outputs.

\section{Conclusion}
\textcolor{black}{We propose the MAC framework to achieve the adaptive collaboration of LLMs. First, MAC employs Pareto-frontier factors analysis to select optimal LLMs as agents, where we demonstrate that the four core metrics of this analysis are statistically independent, enabling a localized deployment that effectively balances hardware efficiency and model robustness. Next, MAC iteratively masks the agent with the lowest Cross-Consistency (CC) value. Such a hard-masking strategy minimizes inconsistent outputs, effectively mitigates overconfident hallucinations, and preserves a robust majority consensus. Finally, the remaining agents propagate their refined outputs to the next layer. Extensive experiments are conducted on closed-ended and open-ended medical benchmarks and the mathematical reasoning dataset, consistently demonstrating the broad generalizability of our framework. Ultimately, MAC provides a highly efficient, deployable, and clinically robust multi-agent solution for real-world decision-making.  Nevertheless, the trade-off between minority-opinion preservation and error suppression deserves further thorough study. Exploring hybrid mechanisms, such as the selective dynamic re-activation of minority opinions based on external clinical guidelines, is a highly promising direction for our future work.}

\ifCLASSOPTIONcaptionsoff
  \newpages
\fi

\balance

\bibliographystyle{IEEEtran}
\bibliography{refs}

\vspace{-0.6cm}
\begin{IEEEbiography}[{\includegraphics[width=1in,height=1.25in, clip,keepaspectratio]{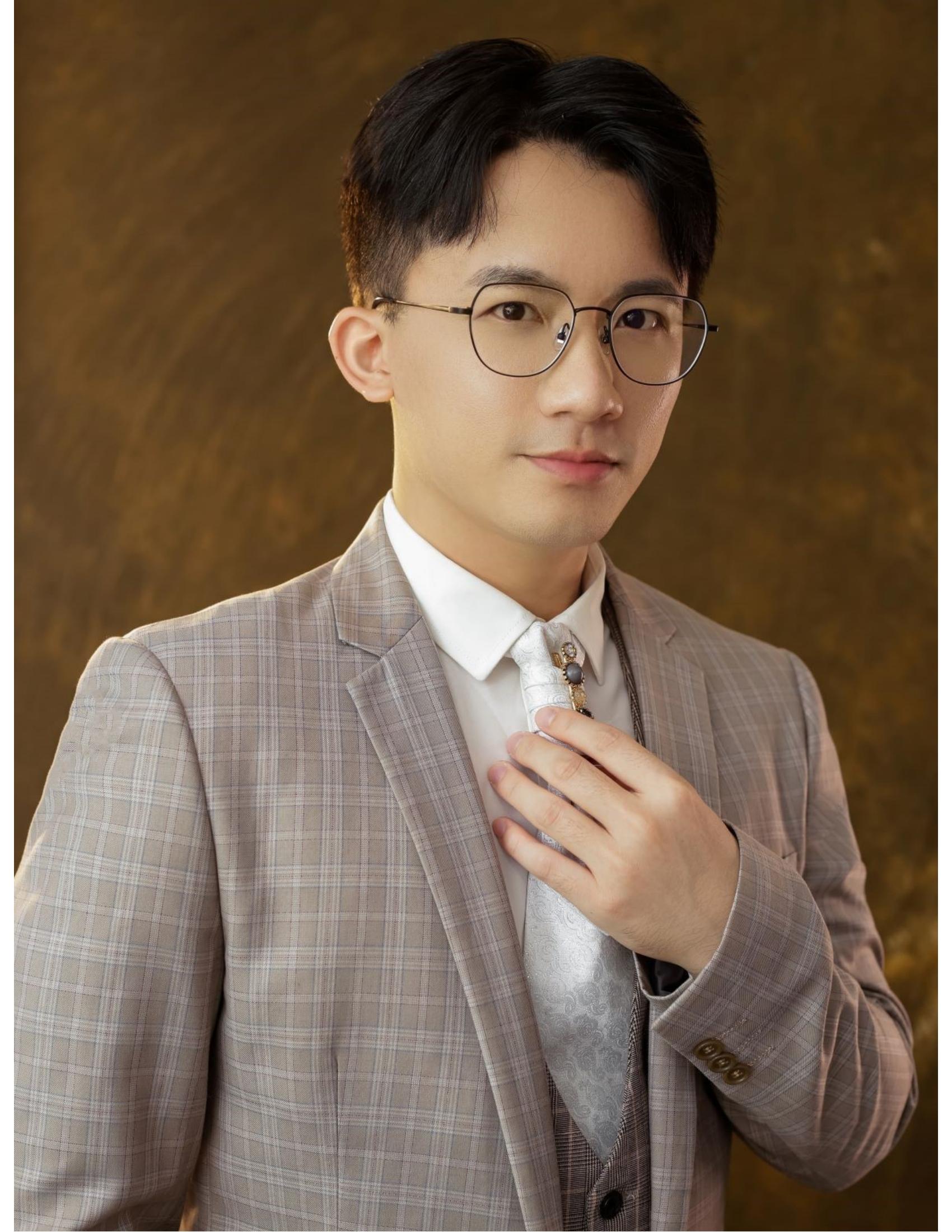}}]{Zhihao Peng}
received the B.S. and M.S. degrees from Guangdong University of Technology in 2016 and 2019, and the Ph.D. degree in Computer Science from the City University of Hong Kong in 2023. He is currently a Postdoctoral Fellow in the Department of Electronic Engineering at The Chinese University of Hong Kong (funded by RTH-ITF).

His research undergoes a paradigm shift toward structured cognition, evolving from explicit geometric topology discovery to the systemic reconstruction of clinical decision logic. His current interests focus on Brain-inspired AI and Medical Multi-agent Systems, including the development of neuroimaging foundation models, collaborative LLM frameworks for complex medical reasoning, and large-scale multimodal benchmarks. He has published over 10 papers in prestigious venues such as Nature Biomedical Engineering, IEEE TIP, TCSVT, CVPR, NeurIPS, ACM MM, and MICCAI.
\end{IEEEbiography}
\vspace{-0.6cm}

\begin{IEEEbiography}[{\includegraphics[width=1in,height=1.25in, clip,keepaspectratio]{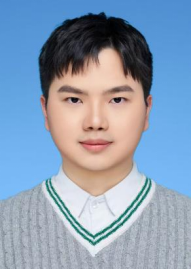}}]{Liuxin Bao} received the B.S. degree from Hangzhou Dianzi University, Hangzhou, China, in 2022. He is currently pursuing the Ph.D. degree with the School of Automation, Hangzhou Dianzi University, Hangzhou, China. He is also working as a research assistant at The Chinese University of Hong Kong. His research interests include saliency detection, semantic segmentation, and medical agent frameworks. He has published over 10 papers in prestigious venues such as IEEE TIP, TCYB, and TITS.
\end{IEEEbiography}

\vspace{-0.6cm}
\begin{IEEEbiography}[{\includegraphics[width=1in,height=1.25in, clip,keepaspectratio]{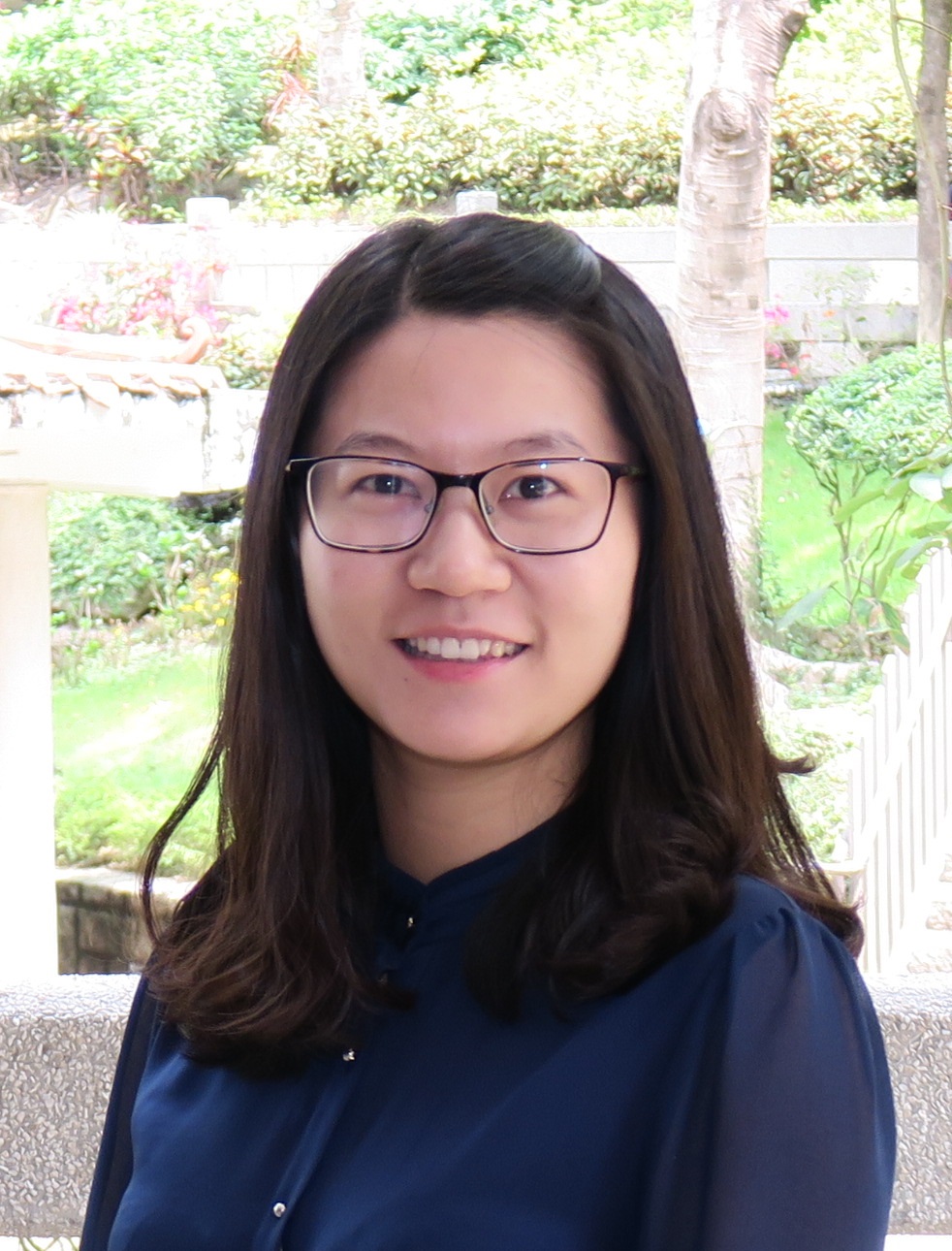}}]
{Yixuan Yuan} received the B.S. degree from the College of Information Countermeasure, Northwestern Polytechnical University, Xi’an, China, in 2010, and the Ph.D. degree from the College of Biomedical Engineering, The Chinese University of Hong Kong, Hong Kong, China, in 2016. From 2018-2022, she was an assistant professor in the Department of Electrical Engineering at the City University of Hong Kong. She is currently an associate professor in the Department of Electronic Engineering at the Chinese University of Hong Kong. Her research interests include medical image analysis and deep learning in healthcare.
\end{IEEEbiography}

\end{document}